\documentclass{article}

\usepackage{PRIMEarxiv}

\usepackage[utf8]{inputenc} % allow utf-8 input
\usepackage[T1]{fontenc}    % use 8-bit T1 fonts
\usepackage{hyperref}       % hyperlinks
\usepackage{url}            % simple URL typesetting
\usepackage{booktabs}       % professional-quality tables
\usepackage{amsfonts}       % blackboard math symbols
\usepackage{nicefrac}       % compact symbols for 1/2, etc.
\usepackage{microtype}      % microtypography
\usepackage{lipsum}
\usepackage{fancyhdr}       % header
\usepackage{graphicx}       % graphics
\graphicspath{{media/}}     % organize your images and other figures under media/ folder
\usepackage{amsmath}
\usepackage{amssymb}
\usepackage{mathtools}
\usepackage{amsthm}
\usepackage{adjustbox}
\usepackage{caption}
\usepackage{subcaption}

\title{MAGPrompt: Message-Adaptive Graph Prompt Tuning for Graph Neural Networks}

\author{
  Long D. Nguyen \\
  School of Mathematics and Statistics \\
  Victoria University of Wellington \\
  Wellington, New Zealand\\
  \texttt{duylong.nguyen@vuw.ac.nz} \\
   \And
  Binh P. Nguyen \\
  School of Mathematics and Statistics \\
  Victoria University of Wellington \\
  Wellington, New Zealand \\
  \texttt{binh.p.nguyen@vuw.ac.nz} \\
}

\begin{document}
\maketitle

\begin{abstract}
Pre-trained graph neural networks (GNNs) transfer well, but adapting them to downstream tasks remains challenging due to mismatches between pre-training objectives and task requirements. Graph prompt tuning offers a parameter-efficient alternative to fine-tuning, yet most methods only modify inputs or representations and leave message passing unchanged, limiting their ability to adapt neighborhood interactions. We propose message-adaptive graph prompt tuning, which injects learnable prompts into the message passing step to reweight incoming neighbor messages and add task-specific prompt vectors during message aggregation, while keeping the backbone GNN frozen. The approach is compatible with common GNN backbones and pre-training strategies, and applicable across downstream settings. Experiments on diverse node- and graph-level datasets show consistent gains over prior graph prompting methods in few-shot settings, while achieving performance competitive with fine-tuning in full-shot regimes.
\end{abstract}

% keywords can be removed

\section{Introduction}

Graphs are a fundamental data structure underlying a wide range of applications, from social and knowledge networks to traffic forecasting and biomedical informatics~\cite{wei2023dual,zhou2023hierarchical,wang2024knowledge,platonov2023critical,cui2020deterrent,nguyen2024smiles}.
Graph neural networks (GNNs) have become a common paradigm for learning on such graph-structured data by propagating and aggregating information through message passing~\cite{kipf2017semisupervised,veličković2018graph,xu2018how,chen2020simple,Hu*2020Strategies}.
Despite their success, conventional GNN pipelines are typically trained end-to-end for a specific task, which raises practical challenges~\cite{sun2022gppt,fang2023universal,fu2025edge}.
High-quality labels are often costly to obtain, and models optimized for one objective can generalize poorly to others, even on the same underlying graph~\cite{wang2024gft}.
To reduce reliance on labels and improve transferability, self-supervised pre-training of GNNs on large-scale unlabeled graphs has emerged as a widely adopted approach~\cite{Hu*2020Strategies,veličković2018deep,you2020graph,xia2022simgrace,nguyen2025molhfcnet}.

The goal of GNN pre-training is to learn representations that transfer across tasks.
However, straightforward fine-tuning can be brittle when there is a mismatch between pre-training objectives and downstream goals.
For example, a model pre-trained with link prediction or attribute masking~\cite{Hu*2020Strategies} may later be adapted to node- or graph-level classification, where the relevant decision boundaries and neighborhood interactions differ substantially.
Such objective discrepancies can lead to sub-optimal adaptation~\cite{fang2023universal} and may induce negative transfer or catastrophic forgetting~\cite{Hu*2020Strategies,fang2023universal,zhou2021overcoming}.
Motivated by these limitations, graph prompt tuning has been proposed as a parameter-efficient alternative: it freezes the pre-trained GNN backbone and introduces a small set of learnable prompts to steer the model toward a downstream task~\cite{fang2023universal,sun2023all,huang2025one,fu2025edge,fu2025graphtop}.

\begin{table}[!ht]
\centering
\caption{Comparison of representative graph prompt tuning methods.
``Reweights Messages?
'' indicates whether a method explicitly assigns adaptive, edge-specific weights to messages during message passing, rather than only augmenting features, representations, or graph structure.}
\label{tab:prompt_message}
\resizebox{0.7\linewidth}{!}{
\begin{tabular}{lcc}
\toprule
\textbf{Method} & \textbf{Prompt Injection Level} & \textbf{Reweights Messages?} \\
\midrule
GPPT~\cite{sun2022gppt}                 & Task Embedding        & $\times$ \\
GraphPrompt~\cite{liu2023graphprompt}   & Readout               & $\times$ \\
GraphPrompt+~\cite{yu2024generalized}   & Hidden Representation & $\times$ \\
ALL-in-One~\cite{sun2023all}            & Node Feature          & $\times$ \\
GPF / GPF-plus~\cite{fang2023universal} & Node Feature          & $\times$ \\
MultiGPrompt~\cite{yu2024multigprompt}  & Hidden Representation & $\times$ \\
UniPrompt~\cite{huang2025one}           & Graph Topology        & $\times$ \\
GraphTOP~\cite{fu2025graphtop}          & Graph Topology        & $\times$ \\
EdgePrompt~\cite{fu2025edge}            & Edge Message Content  & $\times$ \\
\midrule
\textbf{MAGPrompt (Ours)}               & Edge Message Weighting & \textbf{\checkmark} \\
\bottomrule
\end{tabular}
}
\end{table}

As summarized in Table~\ref{tab:prompt_message}, most existing graph prompt tuning methods adapt pre-trained GNNs by conditioning node features~\cite{fang2023universal,sun2023all}, injecting prompts into hidden representations~\cite{yu2024generalized,yu2024multigprompt}, modifying graph-level readouts~\cite{liu2023graphprompt}, or altering graph structure~\cite{fu2025graphtop,huang2025one}.
EdgePrompt additionally augments message content at the edge level~\cite{fu2025edge}, but still leaves per-neighbor contributions during message passing unchanged.
Crucially, these approaches do not directly modulate how individual neighbors are weighted during aggregation.
%, and instead perturb inputs, representations, or graph structure.
This design choice limits prompt tuning when downstream tasks require neighborhood mixing patterns that differ from those induced during pre-training.
Since message passing is the core mechanism through which GNNs propagate and integrate relational information~\cite{kipf2017semisupervised,veličković2018graph,xu2018how,shi2021masked}, enabling task-specific control over message weighting is essential for fully exploiting prompt-based adaptation.

In this work, we propose a \emph{Message-Adaptive Graph Prompt} framework (\textbf{MAGPrompt}) that directly intervenes in the message-passing process of pre-trained GNNs.
Instead of modifying node features or graph topology, MAGPrompt introduces lightweight prompt parameters that (i) \emph{reweight} messages from neighboring nodes and (ii) \emph{inject} additional prompt signals during aggregation.
This design enables task-specific modulation of neighbor contributions while preserving the original graph structure and keeping the backbone parameters frozen.
MAGPrompt is applicable to a broad class of message-passing GNNs with explicit neighborhood aggregation (e.g., GCN and GIN) under a variety of pre-training strategies, allowing the same prompting mechanism to be applied across downstream tasks without redesigning task-specific prompts.
Extensive experiments demonstrate that adapting message passing yields more expressive prompt tuning than representation-only prompting, across both node and graph classification tasks.
%while retaining strong parameter efficiency.

\noindent\textbf{Contributions.} The main contributions of this work are:
\begin{itemize}
    \item We propose {MAGPrompt}, a message-adaptive graph prompt tuning framework that directly modulates neighborhood messages in pre-trained GNNs, rather than only conditioning node/hidden representations or altering graph topology.
    \item We develop a general prompting mechanism that is compatible with common GNN architectures (e.g., GCN and GIN) and multiple pre-training strategies, while keeping backbone parameters frozen.
    \item Comprehensive experiments show that MAGPrompt consistently outperforms prior representation-based graph prompting methods in few-shot and achieves competitive performance compared to full fine-tuning in full-shot settings, under various pre-trained models. %with substantially fewer trainable parameters.
\end{itemize}

%=============================================================================%
\section{Related Work}

\subsection{Graph Pre-training}
% Self-supervised pre-training has become a key approach for learning transferable graph representations from unlabeled data. 
% Early mutual-information-based methods include Deep Graph Infomax (DGI)~\cite{veličković2018deep}, which contrasts local patch representations with a global graph summary, and InfoGraph~\cite{sun2019infograph}, which extends this idea to graph-level objectives by contrasting whole-graph representations with substructure representations across multiple scales. 
% More recent work has been dominated by contrastive learning, where models are trained to maximize agreement between two correlated views of the same graph. 
% GraphCL~\cite{you2020graph} achieves this via graph augmentations, and MolCLR~\cite{wang2022molecular} tailors augmentation policies to molecular graphs. 
% To reduce sensitivity to augmentation design, SimGRACE~\cite{xia2022simgrace} generates correlated views by perturbing model parameters rather than modifying the input graph. 
% Beyond contrastive objectives, predictive and generative pre-training tasks have also been explored, including context prediction and attribute masking~\cite{Hu*2020Strategies}, as well as motif-based objectives that encourage capturing higher-order semantic patterns~\cite{zhang2021motif}. 
% Overall, these methods suggest that diverse pre-training signals can encode complementary structural and semantic information that benefits downstream graph learning.

Self-supervised pre-training has emerged as a key paradigm for learning transferable graph representations from unlabeled data.
Early mutual-information-based methods include Deep Graph Infomax (DGI)~\cite{veličković2018deep}, which contrasts local node representations with a global graph summary, and InfoGraph~\cite{sun2019infograph}, which extends this principle to graph-level objectives by contrasting whole-graph and substructure representations across multiple scales.
Recent work has been dominated by contrastive learning, where models maximize agreement between correlated views of the same graph.
GraphCL~\cite{you2020graph} constructs such views via graph augmentations, while MolCLR~\cite{wang2022molecular} tailors augmentation strategies to molecular graphs.
To reduce reliance on augmentation design, SimGRACE~\cite{xia2022simgrace} generates alternative views by perturbing model parameters instead of the input graph.
Beyond contrastive objectives, predictive and generative pre-training tasks have also been explored, including context prediction, attribute masking~\cite{Hu*2020Strategies}, and motif-based objectives that capture higher-order semantic patterns~\cite{zhang2021motif}.
Together, these approaches demonstrate that diverse pre-training signals encode complementary structural and semantic information beneficial for downstream graph learning.

\subsection{Graph Prompt Learning}

Graph prompt tuning is a parameter-efficient paradigm %by learning a small set of prompt parameters while keeping the backbone frozen. for adapting pre-trained GNNs  
by optimizing a compact set of prompt parameters while freezing the backbone. Existing methods can be broadly categorized by \emph{where} prompts are injected, and most preserve the original message-passing and aggregation mechanisms.
\emph{Node-feature prompting} methods, such as GPF and GPF-plus~\cite{fang2023universal}, inject global or node-specific prompts into input features, achieving input-level adaptation without modifying neighborhood aggregation.
\emph{Edge prompting} approaches, including EdgePrompt and EdgePrompt+~\cite{fu2025edge}, enrich message content via learnable edge prompts, but still aggregate neighbors using the fixed aggregation rule of the pre-trained backbone.
\emph{Topology prompting} methods, such as GraphTOP~\cite{fu2025graphtop} and UniPrompt~\cite{huang2025one}, adapt graph structure through edge rewiring or auxiliary prompt graphs, yet continue to rely on standard message passing.
At the \emph{representation level}, GraphPrompt~\cite{liu2023graphprompt} introduces prompts at the readout layer, while GraphPrompt+ and MultiGPrompt~\cite{yu2024generalized,yu2024multigprompt} apply layer-wise prompts to intermediate representations.
GPPT~\cite{sun2022gppt} aligns downstream node classification with a masked edge prediction pretext by reformulating it as a link prediction task with learnable tokens.

Despite their empirical success, these approaches primarily adapt inputs, representations, or graph structure, and do not explicitly control \emph{how} messages from different neighbors are weighted during aggregation.
This motivates our work: we introduce message-adaptive prompts that modulate per-neighbor contributions during message passing, enabling task-specific neighborhood mixing while keeping the pre-trained backbone parameters fixed.

%=============================================================================%

\section{Preliminaries}

\paragraph{Graphs and Notation.}
We consider an attributed graph $\mathcal{G}=(\mathcal{V},\mathcal{E},\mathbf{X})$, where
$\mathcal{V}=\{v_1,\ldots,v_n\}$ is the node set,
$\mathcal{E}\subseteq \mathcal{V}\times\mathcal{V}$ is the edge set,
and $\mathbf{X}\in\mathbb{R}^{n\times d_x}$ is the node-feature matrix with $\mathbf{x}_i$ denoting the feature of node $v_i$.
The (optionally self-loop augmented) neighborhood of $v_i$ is denoted by $\mathcal{N}(v_i)$.
The graph structure can be represented by an adjacency matrix $\mathbf{A}\in\{0,1\}^{n\times n}$.
Edges may additionally carry attributes $\mathbf{e}_{ij}\in\mathbb{R}^{d_e}$; we denote the collection of edge features by $\mathbf{E}$ and omit it when unavailable.

\paragraph{Message Passing in GNNs.}
%Graph neural networks 
Node representations in GNN are computed via iterative message passing.
Let $\mathbf{h}_i^{(l)}\in\mathbb{R}^{d_l}$ denote the representation of node $v_i$ at layer $l$, with $\mathbf{h}_i^{(0)}=\mathbf{x}_i$.
A generic message-passing layer is defined as
\begin{align}
\mathbf{m}_{ij}^{(l)} &= \textsc{MSG}^{(l)}\!\left(\mathbf{h}_i^{(l-1)}, \mathbf{h}_j^{(l-1)},
\mathbf{e}_{ij}\right),\label{eq:gnn_msg}\\
\bar{\mathbf{m}}_{i}^{(l)} &= \textsc{AGG}^{(l)}\!\left(\{\mathbf{m}_{ij}^{(l)} \mid v_j \in \mathcal{N}(v_i)\}\right),
\label{eq:gnn_agg}\\
\mathbf{h}_i^{(l)} &= \textsc{UPD}^{(l)}\!\left(\mathbf{h}_i^{(l-1)}, \bar{\mathbf{m}}_{i}^{(l)}\right),\label{eq:gnn_udp}
\end{align}
where $\textsc{MSG}(\cdot)$ computes message content, $\textsc{AGG}(\cdot)$ aggregates neighbor messages, and $\textsc{UPD}(\cdot)$ updates node states.
Specific GNN architectures (e.g., GCN, GIN) correspond to different instantiations of these operators. 
% This formulation subsumes common GNN variants, including GCN- and GIN-style message passing as well as attention-based architectures such as GAT.

\paragraph{Pre-training and Fine-tuning.}
In the pre-train--fine-tune paradigm, a GNN encoder $f_{\boldsymbol{\theta}}$ is first trained on unlabeled graphs using self-supervised objectives, and then adapted to a labeled downstream dataset $\mathcal{D}$.
Given a task-specific head $g_{\boldsymbol{\phi}}$, fine-tuning jointly optimizes $(\boldsymbol{\theta},\boldsymbol{\phi})$ by minimizing
\begin{equation}
\min_{\boldsymbol{\theta},\boldsymbol{\phi}}
\frac{1}{|\mathcal{D}|}\!\sum_{(\mathbf{A},\mathbf{X},\mathbf{E},y)\in\mathcal{D}}
\mathcal{L}\!\left(g_{\boldsymbol{\phi}}\!\left(f_{\boldsymbol{\theta}}(\mathbf{A},\mathbf{X},\mathbf{E})\right), y\right),
\end{equation}
where $\mathcal{L}$ is a supervised loss and $\mathbf{E}$ is omitted if edge attributes are unavailable.
While effective, full fine-tuning can be parameter-inefficient and prone to overfitting. % in low-resource regimes.

\paragraph{Graph Prompt Tuning.}
Graph prompt tuning adapts a pre-trained encoder by freezing $\boldsymbol{\theta}$ and learning a small set of prompt parameters $\boldsymbol{\Psi}$ together with a lightweight head $\boldsymbol{\phi}$.
Let $\textsc{Predict}_{\boldsymbol{\Psi}}(\mathbf{A},\mathbf{X},\mathbf{E};f_{\boldsymbol{\theta}})$ denote the prompt-augmented prediction pipeline.
The downstream objective becomes
\begin{equation}
\min_{\boldsymbol{\Psi},\boldsymbol{\phi}}
\frac{1}{|\mathcal{D}|}\!\sum_{(\mathbf{A},\mathbf{X},\mathbf{E},y)\in\mathcal{D}}
\mathcal{L}\!\left(g_{\boldsymbol{\phi}}\!\left(\textsc{Predict}_{\boldsymbol{\Psi}}(\mathbf{A},\mathbf{X},\mathbf{E};f_{\boldsymbol{\theta}})\right), y\right).
\end{equation}
Compared to fine-tuning, prompt tuning substantially reduces trainable parameters while retaining competitive performance.
Most existing methods inject prompts by modifying inputs, hidden representations, or graph structure, while treating the message and aggregation operators as fixed.

\paragraph{Scope of This Work.}
In contrast, we study \emph{message-adaptive prompting}, which intervenes directly in message passing.
Specifically, we introduce learnable prompts that reweight per-neighbor messages and inject additional prompt signals during aggregation, while preserving the original graph topology and keeping the encoder $f_{\boldsymbol{\theta}}$ frozen.

%=============================================================================%

\section{Method}

\begin{figure*}[!ht]
\centering
\begin{subfigure}[b]{\linewidth}
\centering
\includegraphics[width=0.8\linewidth]{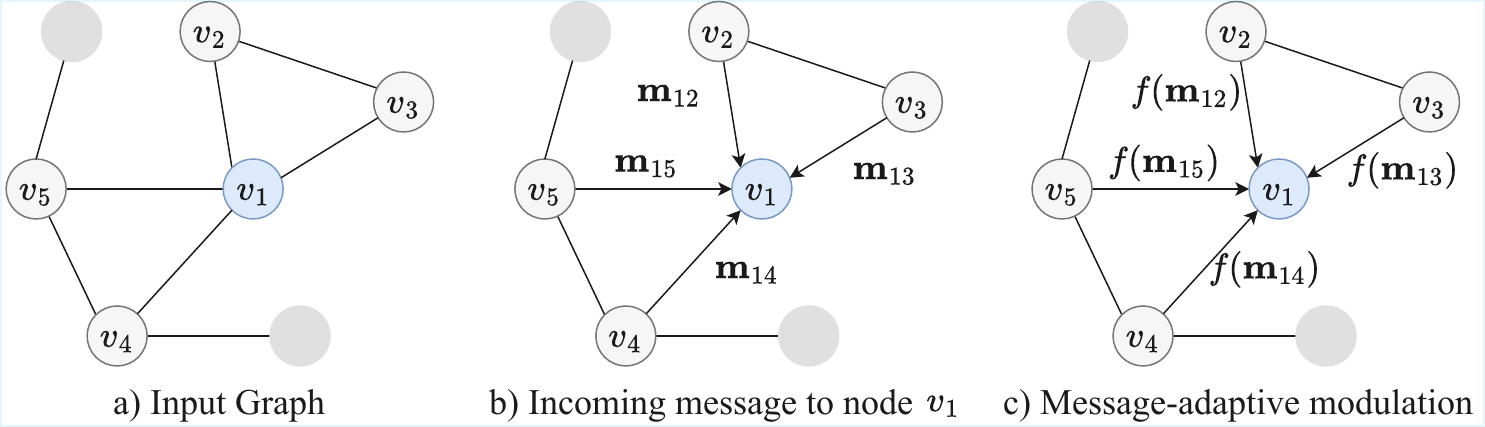}
\end{subfigure}
\caption{Illustration of MAGPrompt. (a) Input graph. (b) Standard message passing aggregates messages from neighbors of $v_1$. (c) MAGPrompt introduces edge-specific message-adaptive modulation $f(\mathbf{m}_{ij}) = a_{ij} \mathbf{m}_{ij} + \mathbf{p}_{ij}$ before aggregation.}
\label{fig:magprompt}
\end{figure*}

\subsection{Overview}

Let $f_{\boldsymbol{\theta}}$ be a pre-trained GNN with $L$ message-passing layers, whose parameters $\boldsymbol{\theta}$ are \emph{frozen} during downstream adaptation.
At layer $l$, the frozen backbone computes edge messages and aggregates them as Eqs.~\eqref{eq:gnn_msg} --~\eqref{eq:gnn_udp}.
Our key idea is to introduce lightweight, trainable prompts that \emph{modulate per-neighbor message contributions during aggregation} by (i) reweighting neighbor contributions and (ii) injecting additive prompt signals, while keeping ${\boldsymbol{\theta}}$ unchanged.
% The gating coefficients and prompt parameters are optimized solely
% to enable task-specific reweighting of neighbor contributions without altering pretrained representations. 
% Figure \ref{fig:magprompt} illustrates the overview of our proposed MAGPrompt method. 
An overview of MAGPrompt is visualized in Figure \ref{fig:magprompt}.

% ============================================================
\subsection{MAGPrompt}
\label{subsec:magprompt}
We propose {Message-Adaptive Graph Prompting (MAGPrompt)}, which intervenes in message passing via a \emph{trainable edge gate} and a \emph{message-wise additive prompt}.
MAGPrompt introduces only a small set of trainable parameters per layer: a prompt vector $\mathbf{p}^{(l)}\in\mathbb{R}^{d_l}$ and a gating module that outputs a scalar weight $a_{ij}^{(l)}$ for each edge $(v_i,v_j)$.

\paragraph{Message-adaptive reweighting.}
At each layer $l$, MAGPrompt first projects node representations into a $d_a$-dimentional gating space:
\begin{equation}
\mathbf{b}_i^{(l)} = \mathbf{h}_i^{(l-1)}\mathbf{W}_{g}^{(l)}  + \mathbf{c}^{(l)};
\mathbf{W}_{g}^{(l)}\in\mathbb{R}^{d_{l-1}\times d_a},\ \mathbf{c}^{(l)}\in\mathbb{R}^{d_a}, 
\label{eq:proj_b}
\end{equation}
where $d_a$ may be viewed as the number of attention heads.
For an edge $(v_i,v_j)$, head-wise attention scores are computed
\begin{equation}
\tilde{\boldsymbol{\alpha}}_{ij}^{(l)} =
\textsc{LeakyReLU}\!\left(
(\mathbf{b}_j^{(l)} \odot \mathbf{w}_s^{(l)})
+
(\mathbf{b}_i^{(l)} \odot \mathbf{w}_d^{(l)})
\right),
\label{eq:node_prj_att}
\end{equation}
where $\mathbf{w}_s^{(l)}, \mathbf{w}_d^{(l)} \in \mathbb{R}^{d_a}$ denote attention vectors, $\odot$ represents the element-wise product, and $\textsc{Softmax}$ is applied independently per head across the neighborhood,
\begin{equation}
\boldsymbol{\alpha}_{ij}^{(l)}[k]
=
\frac{
\exp(\tilde{\boldsymbol{\alpha}}_{ij}^{(l)}[k])
}{
\sum_{j' \in \mathcal{N}(i)}
\exp(\tilde{\boldsymbol{\alpha}}_{ij'}^{(l)}[k])
}.
\label{eq:alpha_softmax}
\end{equation}
The scalar message gate is obtained by averaging over heads,
\begin{equation}
\bar{\alpha}_{ij}^{(l)} = \frac{1}{d_a} \sum_{k=1}^{d_a} \boldsymbol{\alpha}_{ij}^{(l)}[k], \quad a_{ij}^{(l)} = (1-\beta)\,\bar{\alpha}_{ij}^{(l)} + \beta
\label{eq:gate_beta}
\end{equation}
yielding a bounded gate $a_{ij}^{(l)} \in [\beta,1]$ that stabilizes message propagation.

\paragraph{Prompted combination.}
MAGPrompt modulates aggregation by reweighting each incoming message and injecting a message-wise prompt:
\begin{align}
\tilde{\mathbf{m}}_{ij}^{(l)} &= a_{ij}^{(l)}\cdot \mathbf{m}_{ij}^{(l)}+\mathbf{p}^{(l)}, \label{eq:prompted_msg_mag}\\
\bar{\mathbf{m}}_{i}^{(l)} &= \textsc{AGG}^{(l)}\!\left(\left\{ \tilde{\mathbf{m}}_{ij}^{(l)} \mid v_j \in \mathcal{N}(v_i) \right\}\right), \label{eq:prompted_agg_mag}\\
\mathbf{h}_i^{(l)} &= \textsc{UPD}^{(l)}\!\left(\mathbf{h}_i^{(l-1)}, \bar{\mathbf{m}}_i^{(l)}\right). \label{eq:prompted_upd_mag}
\end{align}
This design, which adds a shared prompt vector $\mathbf{p}^{(l)}$ to each message, constitutes MAGPrompt.

\paragraph{Discussion.} 
Unlike EdgePrompt~\cite{fu2025edge}, which injects additive prompts without explicitly reweighting neighbor contributions, MAGPrompt directly modulates effective aggregation weights via $a_{ij}^{(l)}$, enabling task-specific control over neighborhood interactions while keeping the backbone parameters $\boldsymbol{\theta}$ fixed.
For attention-based backbones (e.g., GAT), MAGPrompt can be viewed as an external, task-specific residual modulation applied on top of the learned attention weights, complementing rather than replacing the backbone attention mechanism.

The impact of additive prompts depends on the aggregation operator, which remained fixed.
% Under sum-based aggregation (e.g., GIN), prompt contributions scale with node degree, whereas mean-based aggregation (e.g., GCN) yields degree-normalized effects.
MAGPrompt does not impose a specific aggregation scheme; instead, the learned gating coefficients adaptively regulate message contributions, mitigating excessive prompt influence across different backbone designs.

% ============================================================
\subsection{MAGPrompt+}
\label{subsec:magprompt+}

While MAGPrompt uses a single message-wise prompt vector, complex graphs may require multiple prompt patterns to capture heterogeneous relational semantics.
We propose {MAGPrompt+}, which \emph{inherits the same message-adaptive gating} $a_{ij}^{(l)}$ from MAGPrompt and extends the additive prompt to an \emph{edge-adaptive composition} from a learnable prompt basis.
At layer $l$, MAGPrompt+ maintains $M_l$ prompt basis vectors
$\mathcal{P}^{(l)}=\{\mathbf{p}_1^{(l)},\dots,\mathbf{p}_{M_l}^{(l)}\}$ with $\mathbf{p}_m^{(l)}\in\mathbb{R}^{d_l}$.
For each edge $(v_i,v_j)$, mixture logits are computed using matrix $\mathbf{W}_b^{(l)}\in \mathbb{R}^{d_a \times M_l }$, bias vector $\mathbf{d}^{(l)} \in \mathbb{R}^{M_l}$ and the projected features $\mathbf{b}_i^{(l)}$ and $\mathbf{b}_j^{(l)}$:
\begin{equation}
\begin{aligned}
\mathbf{z}_{ij}^{(l)} &=
\textsc{LeakyReLU}\!\left(
\left[\mathbf{b}_i^{(l)} +  \mathbf{b}_j^{(l)}\right]\mathbf{W}_b^{(l)} + \mathbf{d}^{(l)}
\right),\\
% \mathbf{W}_b^{(l)} &\in \mathbb{R}^{d_a \times M_l }, \qquad
% \mathbf{d}^{(l)} \in \mathbb{R}^{M_l},
\end{aligned}
\label{eq:mix_logits}
\end{equation}
and apply a softmax to obtain mixture weights:
\begin{equation}
\boldsymbol{\pi}_{ij}^{(l)}=\textsc{Softmax}\!\left(\mathbf{z}_{ij}^{(l)}\right),
\qquad
\sum_{m=1}^{M_l}\pi_{ijm}^{(l)}=1.
\label{eq:mix_weights}
\end{equation}
The edge-adaptive prompt is the combination
\begin{equation}
\mathbf{p}_{ij}^{(l)}=\sum_{m=1}^{M_l}\pi_{ijm}^{(l)}\,\mathbf{p}_m^{(l)}.
\label{eq:edge_prompt_comp}
\end{equation}
MAGPrompt+ then injects $\mathbf{p}_{ij}^{(l)}$ at the \emph{message level} while still reweighting the frozen backbone message:
\begin{align}
\tilde{\mathbf{m}}_{ij}^{(l)} &= a_{ij}^{(l)}\cdot \mathbf{m}_{ij}^{(l)} + \mathbf{p}_{ij}^{(l)}, \label{eq:prompted_msg_plus}\\
\bar{\mathbf{m}}_{i}^{(l)} &= \textsc{AGG}^{(l)}\!\left(\left\{ \tilde{\mathbf{m}}_{ij}^{(l)} \mid v_j \in \mathcal{N}(v_i) \right\}\right), \label{eq:prompted_agg_plus}\\
\mathbf{h}_i^{(l)} &= \textsc{UPD}^{(l)}\!\left(\mathbf{h}_i^{(l-1)}, \bar{\mathbf{m}}_i^{(l)}\right). \label{eq:prompted_upd_plus}
\end{align}
Compared to MAGPrompt, MAGPrompt+ offers finer-grained, edge-specific modulation of message content while preserving parameter efficiency via a small prompt basis.

\textbf{Proposition 1.} \textit{(Structure Preservation and Equivariance)}
\label{para:propo_1}
MAGPrompt+ preserves the underlying graph structure and the permutation equivariance/invariance properties of the pre-trained GNN.
Specifically, MAGPrompt+ does not alter the edge set of the input graph and maintains the equivariance of node representations and the invariance of graph-level representations under node permutations.

The detailed proof for Proposition~1 is provided in Appendix~\ref{app:proof_1}.

% ============================================================
\subsection{Prompt-collapse Regularization}
A practical issue in MAGPrompt+ is \emph{prompt collapse}: mixture weights may concentrate on only a few basis prompts across most edges, effectively reducing the diversity of learned prompts.
To encourage balanced utilization, we regularize the aggregate prompt usage within a mini-batch.

% For node classification with mini-batching, $E_B$ denotes the set of edges in the induced computational subgraph corresponding to the mini-batch, including edges introduced by neighborhood sampling.
% Let $\mathcal{E}_{\mathcal{B}}$ be the set of edges appearing in a mini-batch $\mathcal{B}$ and $\boldsymbol{\pi}_{ij}^{(l)}\in\mathbb{R}^{M_l}$ be the edge-wise mixture weights from Eq.~\eqref{eq:mix_weights}.
Let $\mathcal{E}_{\mathcal{B}}$ denote the set of edges in a mini-batch $\mathcal{B}$; for node classification, this corresponds to the edges of the induced computational subgraph, including those introduced by neighborhood sampling.
Let $\boldsymbol{\pi}_{ij}^{(l)} \in \mathbb{R}^{M_l}$ be the edge-wise mixture weights from~\eqref{eq:mix_weights}.
We define the usage vector $\mathbf{s}^{(l)}\in\mathbb{R}^{M_l}$:
\begin{equation}
s_m^{(l)}=\sum_{(i,j)\in\mathcal{E}_{\mathcal{B}}}\pi_{ijm}^{(l)} ,
\qquad m=1,\dots,M_l.
\label{eq:usage_vector}
\end{equation}
Prompt collapse corresponds to an imbalanced $\mathbf{s}^{(l)}$.
We penalize the squared coefficient of variation of $\mathbf{s}^{(l)}$:
\begin{equation}
\mathcal{L}_{\mathrm{pc}}^{(l)}
=
\frac{1}{M_l}\sum_{m=1}^{M_l}\frac{\left(s_m^{(l)}-\bar{s}^{(l)}\right)^2}{\left(\bar{s}^{(l)}\right)^2+\epsilon},
\qquad
\bar{s}^{(l)}=\frac{1}{M_l}\sum_{m=1}^{M_l}s_m^{(l)},
\label{eq:pc_loss}
\end{equation}
where $\epsilon>0$ is a small constant for numerical stability.
Minimizing $\mathcal{L}_{\mathrm{pc}}^{(l)}$ encourages more uniform prompt usage without removing edge-wise adaptivity.
The overall training objective for MAGPrompt+ is
\begin{equation}
\mathcal{L}
=
\mathcal{L}_{\mathrm{task}}
+
\lambda_{\mathrm{pc}}\frac{1}{L}
\sum_{l=1}^{L}\mathcal{L}_{\mathrm{pc}}^{(l)},
\label{eq:full_objective}
\end{equation}
where $\mathcal{L}_{\mathrm{task}}$ is the downstream supervised loss (e.g., cross-entropy) and $\lambda_{\mathrm{pc}}$ controls the regularization strength.

\textbf{Proposition 2.} \textit{(Balanced Prompt Utilization)}
\label{para:propo_2}
The prompt-collapse regularization in MAGPrompt+ encourages balanced utilization of the prompt basis vectors by penalizing uneven prompt assignment across edges.

The detailed proof for Proposition~2 is provided in Appendix~\ref{app:proof_2}.

% ============================================================
\subsection{Computational Complexity}
% Let $n=|\mathcal{V}|$ and $m=|\mathcal{E}|$ denote the numbers of nodes and edges, respectively, and let $d$ be the hidden dimension.
% For common backbones (e.g., GCN/GIN/GINE), one message-passing layer typically costs $\mathcal{O}(|\mathcal{E}|\,d + |\mathcal{V}|\,d^2)$.

% \textbf{MAGPrompt.}
% MAGPrompt adds (i) a node projection in Eq.~\eqref{eq:proj_b} with cost $\mathcal{O}(n\,d\,d_a)$ and
% (ii) head-wise attention scoring and neighborhood-wise softmax in Eqs.~\eqref{eq:node_prj_att}--\eqref{eq:gate_beta} with cost $\mathcal{O}(n\,d_a + m\,d_a)$ per layer.
% Prompted message scaling and injection in Eq.~\eqref{eq:prompted_msg_mag} is element-wise and costs $\mathcal{O}(m\,d)$, matching the backbone edge term.
% When $d_a \ll d$ and treated as a small constant, the overall asymptotic complexity remains
% $\mathcal{O}(m\,d + n\,d^2)$, identical to that of the frozen backbone GNN.

% \textbf{MAGPrompt+.}
% MAGPrompt+ retains the same gating cost as MAGPrompt and additionally computes an edge-adaptive prompt by mixing $M_l$ basis vectors.
% For each edge, mixture computation Eq.~\eqref{eq:mix_logits} and prompt composition Eq.~\eqref{eq:edge_prompt_comp} incur $\mathcal{O}(M_l(d+d_a))$ cost, yielding an additional $\mathcal{O}(m\,M_l(d+d_a))$ term per layer.
% Since $M_l$ is chosen as a small constant in practice, MAGPrompt+ remains linear in the number of edges and preserves the scalability of the backbone.

MAGPrompt introduces lightweight, message-level adaptations on top of a frozen GNN backbone.
When the attention head dimension $d_a$ and prompt basis size $M_l$ are treated as small constants, the additional computation and parameters introduced by MAGPrompt and MAGPrompt+ do not change the asymptotic time or memory complexity of the backbone.
Both variants remain linear in the number of nodes and edges, and are more parameter-efficient than fine-tuning.
% the full model. 
A detailed analysis is provided in Appendix~\ref{app:complexity}.

%=============================================================================%

\section{Experiments}
\label{sec:experiments}

% \textcolor{red}{We evaluate} \textbf{MAGPrompt} and \textbf{MAGPrompt+} on node-level and graph-level downstream tasks to answer:
% \textbf{(Q1)} Does message-adaptive prompting improve few-shot and full-shot adaptation over existing graph prompting baselines?
% \textbf{(Q2)} Does MAGPrompt+ bring additional gains via compositional prompts?
% \textbf{(Q3)} How sensitive is MAGPrompt+ to the prompt basis size and prompt-collapse regularization?

% We evaluate \textbf{MAGPrompt} and \textbf{MAGPrompt+} on node- and graph-level tasks to study:
% \textbf{(Q1)} whether message-adaptive reweighting of neighbor contributions improves adaptation over existing graph prompting methods under few-shot and full-shot settings;
% \textbf{(Q2)} what is the impact of the key components of \textbf{MAGPrompt+}, including compositional edge prompts and prompt-collapse regularization, on downstream performance; and
% \textbf{(Q3)} How do hyperparameters such as the prompt basis size, residual gate parameter, and regularization strength influence the expressiveness of message-adaptive prompting?

We evaluate {MAGPrompt} and {MAGPrompt+} on node- and graph-level benchmarks to examine:
\textbf{(Q1)} whether message-adaptive prompt improves downstream adaptation under few-shot and full-shot settings;
\textbf{(Q2)} how the key components of {MAGPrompt+}, including message-adaptive reweighting and compositional edge prompts  contribute to performance; and
\textbf{(Q3)} how hyperparameters such as the prompt basis size, residual gate parameter, regularization strength, and attention hidden dimension influence the model performance.

% ============================================================

\subsection{Datasets and Tasks}
\label{subsec:data_tasks}
We evaluate our methods on both node- and graph-level benchmarks.
For node classification, we use Cora, CiteSeer, Pubmed~\cite{yang2016revisiting}, ogbn-arxiv~\cite{hu2020open}, and Flickr~\cite{graphsainticlr20}.
For graph classification, we adopt five TUDataset~\cite{morris2020tudataset} benchmarks, including ENZYMES, DD, NCI1, NCI109, and Mutagenicity under few-shot settings.
In addition, we evaluate on four MoleculeNet~\cite{wu2018moleculenet} datasets, which contain BACE, BBBP, SIDER, and ClinTox in the standard full-shot setting, following the data curation and scaffold split of~\cite{nguyen2024smiles}.
MoleculeNet datasets include edge attributes, while the remaining benchmarks contain only node features and graph connectivity.
Dataset statistics and preprocessing details are described in Appendix~\ref{app:data}. %~\textcolor{red}{X}.

% ============================================================

\subsection{Pre-training Strategies}
\label{subsec:pretraining}
To assess robustness across pre-training objectives, we consider eight representative strategies:
contrastive methods GraphCL~\cite{you2020graph}, SimGRACE~\cite{xia2022simgrace}, and DGI~\cite{veličković2018deep}, as well as predictive methods LP-GPPT~\cite{sun2022gppt}, LP-GraphPrompt~\cite{liu2023graphprompt}, EdgePred~\cite{kipf2017semisupervised}, AttrMasking, and ContextPred~\cite{Hu*2020Strategies}.
Detailed descriptions are provided in Appendix~\ref{app:pretrain}. %~\textcolor{red}{X}.

% ============================================================

\subsection{Baselines}
\label{subsec:baselines}

We compare MAGPrompt and MAGPrompt+ with state-of-the-art graph prompt tuning methods, including
GPPT~\cite{sun2022gppt}, GraphPrompt~\cite{liu2023graphprompt}, ALL-in-One~\cite{sun2023all}, GPF/GPF-plus~\cite{fang2023universal},
EdgePrompt/EdgePrompt+~\cite{fu2025edge}, and GraphTOP~\cite{fu2025graphtop}.
As GPPT and GraphTOP are primarily designed for node-level tasks, their results are reported only for node classification.
We also include a Linear Probe baseline, which trains a linear classifier on frozen features. %representations without prompts.

% ============================================================

\subsection{Implementation Details}
\label{subsec:impl}

For node classification, we use a 2-layer GCN as the frozen backbone, and a 5-layer GIN-style backbone (GIN for TUDataset and GINE for MoleculeNet datasets) for graph classification, following EdgePrompt~\cite{fu2025edge} and GPF~\cite{fang2023universal} settings.
The hidden dimension is set to 128 (300 for MoleculeNet), with a linear classifier as the downstream head.
We optimize prompt parameters and the classifier using Adam with learning rate $10^{-3}$ and batch size 32 for 200 epochs (100 for MoleculeNet), and fix the gate residual parameter to $\beta=0.5$.
Following~\cite{fu2025edge}, we adopt 5-shot and 50-shot protocols for node and graph classification on TUDataset benchmarks, respectively.
For MoleculeNet, we follow the full-shot evaluation protocol of~\cite{fang2023universal} with scaffold splits~\cite{nguyen2024smiles}.
Unless stated otherwise, we set $d_a=16$, use $M_l=20$ for graph classification and $M_l=10$ for node classification by default, and tune $\lambda_{\mathrm{pc}} \in \{0,0.01,0.1,1.0\}$.
%All experiments are repeated five times with different random seeds, reporting mean$\pm$standard deviation.
All results are averaged over five runs with different random seeds, reporting mean$\pm$standard deviation.

% ============================================================
\subsection{Node Classification Performance}
\label{subsec:node_results}

Table~\ref{tab:node_class} reports the average accuracy (\%) on 5-shot node classification across five benchmark datasets under different pre-training strategies.
Overall, MAGPrompt+ achieves the highest average performance across all settings, consistently outperforming existing graph prompt tuning methods such as GPF-plus, EdgePrompt+, and GraphTOP by a notable margin.
% While GraphTOP attains strong results on certain datasets, its performance degrades substantially on CiteSeer and ogbn-arxiv, indicating limited robustness across tasks.
These results suggest that compositional prompts provide greater expressive capacity than a single shared prompt.
Furthermore, compared with prompt-based approaches that modify node features, message content, or graph topology, message-adaptive reweighting yields more stable and consistent improvements across datasets and pre-training methods.

\begin{table*}[!ht]
\centering
\caption{Average classification Accuracy (\%) on 5-shot node classification tasks across five datasets. The best and second-best results are highlighted in bold and underlined, respectively. The full table can be found in Table~\ref{tab:node_class_full}.}
\label{tab:node_class}
\begin{adjustbox}{width=0.83\textwidth}
\begin{tabular}{l|c|ccccc|c}
\toprule
Pre-training & Tuning & Cora & CiteSeer & Pubmed & ogbn-arxiv & Flickr & Avg. \\
\midrule
 & Linear Probe & 53.05$\pm$4.76 & 38.62$\pm$3.43 & 64.28$\pm$4.51 & 21.15$\pm$1.64 & 24.32$\pm$2.93 & 40.28 \\
 & GPF & 58.52$\pm$4.07 & 43.55$\pm$2.80 & \underline{67.67$\pm$3.14} & 21.73$\pm$1.75 & 23.98$\pm$1.71 & 43.09 \\
 & GPF-plus & 52.24$\pm$4.59 & 38.47$\pm$3.27 & 64.30$\pm$4.58 & 21.03$\pm$1.96 & 25.32$\pm$2.02 & 40.27 \\
GraphCL & EdgePrompt & 58.60$\pm$4.46 & 43.31$\pm$3.23 & \textbf{67.76$\pm$3.01} & 21.90$\pm$1.71 & 24.83$\pm$2.78 & 43.28 \\
 & EdgePrompt+ & \underline{62.88$\pm$6.43} & \underline{46.20$\pm$0.99} & 67.41$\pm$5.25 & \underline{23.18$\pm$1.26} & \underline{25.57$\pm$3.04} & \underline{45.05} \\
 & GraphTOP & 48.35$\pm$4.35 & 40.98$\pm$3.93 & 63.60$\pm$2.60 & 23.14$\pm$2.29 & 25.48$\pm$2.36 & 40.31 \\
 & MAGPrompt & 59.10$\pm$4.81 & 42.87$\pm$3.27 & 64.54$\pm$4.81 & 22.21$\pm$2.52 & 22.96$\pm$0.90 & 42.34 \\
 & MAGPrompt+ & \textbf{66.53$\pm$8.08} & \textbf{50.93$\pm$4.24} & 67.22$\pm$4.35 & \textbf{23.74$\pm$1.72} & \textbf{25.85$\pm$3.29} & \textbf{46.85} \\

 \midrule
 
 & Linear Probe & 52.27$\pm$2.74 & 40.45$\pm$3.55 & 56.72$\pm$3.80 & 20.75$\pm$2.92 & 25.53$\pm$3.98 & 39.14 \\

 & GPF & 58.23$\pm$4.19 & 44.87$\pm$4.35 & 61.55$\pm$3.79 & 21.86$\pm$2.91 & 26.51$\pm$4.69 & 42.60 \\
 & GPF-plus & 52.27$\pm$3.34 & 41.02$\pm$3.49 & 56.95$\pm$3.86 & 21.44$\pm$3.77 & 28.35$\pm$5.50 & 40.01 \\
SimGRACE & EdgePrompt & 58.37$\pm$4.51 & 43.94$\pm$4.15 & 61.10$\pm$3.69 & 21.85$\pm$2.54 & \textbf{30.12$\pm$5.04} & 43.08 \\
 & EdgePrompt+ & \underline{62.40$\pm$7.97} & \underline{46.62$\pm$2.53} & \underline{64.91$\pm$5.58} & \underline{22.74$\pm$2.34} & \underline{28.50$\pm$4.08} & \underline{45.03} \\
 & GraphTOP & 49.06$\pm$3.64 & 39.02$\pm$7.97 & 55.33$\pm$3.71 & 22.50$\pm$2.06 & 26.12$\pm$3.12 & 38.41 \\
 & MAGPrompt & 57.28$\pm$3.86 & 45.88$\pm$5.21 & 59.68$\pm$5.58 & 21.50$\pm$2.55 & 25.10$\pm$3.40 & 41.89 \\
 & MAGPrompt+ & \textbf{65.45$\pm$8.34} & \textbf{51.98$\pm$3.40} & \textbf{66.60$\pm$7.60} & \textbf{23.41$\pm$2.62} & 26.57$\pm$4.89 & \textbf{46.80} \\

\bottomrule
\end{tabular}
\end{adjustbox}
\end{table*}

% ============================================================

\subsection{Graph Classification Performance}
\label{subsec:graph_results}

% \paragraph{TUDataset (50-shot).}
% Table~\ref{tab:graph_class} reports average accuracy (\%) on 50-shot graph classification tasks over five TUDataset benchmarks.
% MAGPrompt+ achieves the best average performance under different pre-training strategies, outperforming strong prompt tuning baselines such as EdgePrompt+ and GPF-plus.
% This indicates that composing prompts and adapting aggregation jointly provides a more expressive yet still parameter-efficient mechanism for graph-level adaptation. 

% \paragraph{MoleculeNet (full-shot).}
% Table~\ref{tab:molecule} reports ROC-AUC (\%) on four MoleculeNet benchmarks under full-shot supervision.
% Across both AttrMasking and EdgePred pre-training strategies, \textbf{MAGPrompt+} achieves the highest average performance, consistently outperforming fine-tuning and existing prompt-based baselines.
% Compared with EdgePrompt+ and GPF-plus, which only enrich node or message content, MAGPrompt+ further improves performance by adaptively modulating neighbor contributions, demonstrating strong compatibility with molecular pre-training objectives and robustness across different datasets.

\textbf{TUDataset (50-shot).}
As shown in Table~\ref{tab:graph_class}, MAGPrompt+ achieves the highest average accuracy across all five TUDataset benchmarks and pre-training strategies, outperforming competitive methods such as EdgePrompt+ and GPF-plus. This highlights the effectiveness of jointly composing edge prompts and adapting neighbor contributions for few-shot graph classification.

\begin{table*}[!ht]
\centering
\caption{Average classification Accuracy (\%) on 50-shot graph classification tasks across TUDataset. The best and second-best results are highlighted in bold and underlined, respectively. The full table can be found in Table~\ref{tab:graph_class_full}.}
\label{tab:graph_class}
\begin{adjustbox}{width=0.83\textwidth}
\begin{tabular}{l|c|ccccc|c}
\toprule
Pre-training & Tuning & ENZYMES & DD & NCI1 & NCI109 & Mutagenicity & Avg. \\
\midrule
 & Linear Probe & 30.50$\pm$1.16 & 62.89$\pm$2.19 & 62.49$\pm$1.95 & 61.68$\pm$0.93 & 66.62$\pm$1.87 & 56.84 \\
 & GraphPrompt & 27.83$\pm$1.61 & 64.33$\pm$1.79 & 63.19$\pm$1.71 & 62.18$\pm$0.48 & \underline{67.62$\pm$0.65} & 57.03 \\
 & ALL-in-one & 25.92$\pm$0.55 & 66.54$\pm$1.82 & 57.52$\pm$2.61 & 62.74$\pm$0.78 & 63.43$\pm$2.53 & 55.23 \\
 & GPF & 30.08$\pm$1.25 & 64.54$\pm$2.22 & 62.66$\pm$1.83 & 62.29$\pm$0.90 & 66.54$\pm$1.85 & 57.22 \\
GraphCL & GPF-plus & 31.00$\pm$1.50 & 67.26$\pm$2.29 & 64.56$\pm$1.10 & 62.84$\pm$0.22 & 66.82$\pm$1.63 & 58.50 \\
 & EdgePrompt & 29.50$\pm$1.57 & 64.16$\pm$2.13 & 63.05$\pm$2.11 & 62.59$\pm$0.93 & 66.87$\pm$1.88 & 57.23 \\
 & EdgePrompt+ & \underline{34.00$\pm$1.25} & \underline{67.98$\pm$2.05} & \underline{66.30$\pm$2.54} & \underline{66.52$\pm$0.91} & 67.47$\pm$2.37 & \underline{60.45} \\
 & MAGPrompt & 31.72$\pm$2.53 & 65.49$\pm$2.64 & 63.62$\pm$2.69 & 62.96$\pm$2.09 & 67.56$\pm$1.98 & 58.27 \\
 & MAGPrompt+ & \textbf{36.97$\pm$3.47} & \textbf{69.40$\pm$2.56} & \textbf{67.28$\pm$1.69} & \textbf{67.04$\pm$0.98} & \textbf{68.64$\pm$1.45} & \textbf{61.87} \\

 \midrule
 
 & Linear Probe & 29.08$\pm$1.35 & 62.12$\pm$2.82 & 56.85$\pm$4.35 & 62.27$\pm$0.78 & 66.30$\pm$1.78 & 55.32 \\
 & GraphPrompt & 26.67$\pm$1.60 & 61.61$\pm$1.91 & 58.77$\pm$0.97 & 62.16$\pm$0.89 & 66.37$\pm$1.17 & 55.12 \\
 & ALL-in-one & 24.92$\pm$1.33 & 63.61$\pm$2.12 & 59.14$\pm$2.12 & 59.70$\pm$1.37 & 64.86$\pm$1.60 & 54.45 \\
 & GPF & 28.33$\pm$1.73 & 63.48$\pm$2.08 & 58.14$\pm$4.16 & 62.52$\pm$1.39 & 66.10$\pm$0.96 & 55.71 \\
LP-GPPT & GPF-plus & 29.25$\pm$1.30 & \underline{66.92$\pm$2.34} & 62.93$\pm$3.23 & 64.13$\pm$1.42 & 67.57$\pm$1.45 & 58.16 \\
 & EdgePrompt & 28.33$\pm$3.41 & 64.03$\pm$2.26 & 59.85$\pm$3.15 & 62.98$\pm$1.44 & 66.36$\pm$1.22 & 56.31 \\
 & EdgePrompt+ & \underline{32.75$\pm$2.26} & 66.16$\pm$1.60 & \underline{63.58$\pm$2.07} & \underline{65.15$\pm$1.60} & \underline{68.35$\pm$1.57} & \underline{59.20} \\
 & MAGPrompt & 29.61$\pm$3.13 & 65.95$\pm$1.81 & 61.48$\pm$3.21 & 63.70$\pm$1.67 & 67.57$\pm$1.25 & 57.66 \\
 & MAGPrompt+ & \textbf{37.19$\pm$2.79} & \textbf{69.13$\pm$1.81} & \textbf{65.36$\pm$2.33} & \textbf{66.17$\pm$1.96} & \textbf{70.29$\pm$0.56} & \textbf{61.63} \\

\bottomrule
\end{tabular}
\end{adjustbox}
\end{table*}

\textbf{MoleculeNet (full-shot).}
Table~\ref{tab:molecule} reports ROC-AUC scores on four MoleculeNet benchmarks. Under both AttrMasking and EdgePred pre-training, MAGPrompt+ attains the best average performance, consistently surpassing fine-tuning and existing prompt-based methods. Compared to approaches that only enrich node or message content, MAGPrompt+ further improves performance by adaptively modulating neighbor contributions, demonstrating strong compatibility with molecular pre-training objectives and 
%robustness across 
datasets.

\begin{table*}[!ht]
\centering
\caption{Average classification ROC-AUC (\%) on full-shot graph classification tasks across MoleculeNet datasets. The best and second-best results are highlighted in bold and underlined, respectively. The full table can be found in Table~\ref{tab:molecule_full}.}
\label{tab:molecule}
\begin{adjustbox}{width=0.73\textwidth}
\begin{tabular}{l|c|cccc|c}
\toprule
Pre-training & Tuning & BACE & BBBP & ClinTox & SIDER & Avg. \\
\midrule
 & Fine-tune & 86.58$\pm$4.56 & 91.77$\pm$4.10 & 64.99$\pm$13.10 & 60.05$\pm$4.40 & 75.85 \\
 & GPF & 83.69$\pm$5.51 & 92.73$\pm$2.82 & 68.28$\pm$12.89 & 60.28$\pm$5.36 & 76.25 \\
 & GPF-plus & 84.12$\pm$5.14 & 93.19$\pm$2.06 & 68.28$\pm$12.80 & 60.75$\pm$4.72 & 76.59 \\
AttrMasking & EdgePrompt & 85.51$\pm$5.71 & 92.14$\pm$2.98 & 64.92$\pm$16.18 & 60.16$\pm$5.06 & 75.68 \\
 & EdgePrompt+ & 86.06$\pm$5.54 & 92.79$\pm$3.57 & 66.06$\pm$15.17 & 60.30$\pm$4.97 & 76.30 \\
 & MAGPrompt & \underline{86.79$\pm$4.95} & \underline{93.20$\pm$2.18} & \underline{68.45$\pm$14.39} & \underline{60.84$\pm$5.39} & \underline{77.32} \\
 & MAGPrompt+ & \textbf{86.82$\pm$5.21} & \textbf{94.32$\pm$2.38} & \textbf{71.81$\pm$11.58} & \textbf{62.01$\pm$4.36} & \textbf{78.74} \\

 \midrule
 
 & Fine-tune & 84.67$\pm$4.76 & \underline{92.99$\pm$3.71} & 64.55$\pm$9.48 & \textbf{62.17$\pm$4.27} & 76.09 \\
 & GPF & 83.00$\pm$5.91 & 91.72$\pm$3.71 & 66.11$\pm$9.21 & 59.12$\pm$6.61 & 74.99 \\
 & GPF-plus & 85.10$\pm$5.32 & 91.93$\pm$3.52 & \underline{66.60$\pm$6.39} & 60.05$\pm$5.39 & 75.92 \\
EdgePred & EdgePrompt & 84.32$\pm$5.77 & 92.60$\pm$3.02 & 63.06$\pm$10.81 & 60.10$\pm$6.90 & 75.02 \\
 & EdgePrompt+ & 85.00$\pm$5.58 & 92.68$\pm$3.14 & 65.15$\pm$13.44 & 60.76$\pm$4.95 & 75.90 \\
 & MAGPrompt & \underline{85.66$\pm$5.27} & 92.45$\pm$3.11 & 65.34$\pm$10.61 & 61.23$\pm$6.00 & \underline{76.17} \\
 & MAGPrompt+ & \textbf{86.18$\pm$4.22} & \textbf{93.48$\pm$2.62} & \textbf{66.90$\pm$11.95} & \underline{61.26$\pm$6.12} & \textbf{76.96} \\
\bottomrule
\end{tabular}
\end{adjustbox}
\end{table*}

% ============================================================
\subsection{Model Analysis}
\label{subsec:model_analysis}

\paragraph{Effect of Message-Adaptive Reweighting and Compositional Edge Prompts.}
Table~\ref{tab:each_component} presents an ablation study evaluating the contributions of message-adaptive reweighting and the compositional edge prompts under different graph pre-training strategies. Across all pre-training methods, enabling message reweighting consistently improves performance, indicating that adapting neighbor contribution is critical for downstream tasks. Moreover, incorporating message-wise prompt further enhances performance, particularly when reweighting is applied.  % effective
%The best results are achieved when both components are enabled, demonstrating their complementary effects. 
Notably, these trends hold across diverse pre-training schemes, highlighting the robustness and general applicability of the approach.

\begin{table}[!ht]
\centering
\caption{Average classification Accuracy (\%) on 50-shot graph classification tasks on ENZYMES, NCI1, and NCI109 for different components of MAGPrompt+. RW and EP denote message-adaptive reweighting and compositional edge prompts, respectively.
Best results are in bold; second-best are underlined.}
\label{tab:each_component}
\begin{adjustbox}{width=0.6\textwidth}
\begin{tabular}{l|cc|ccc|c}
\toprule
Pre-training & RW & EP & ENZYMES & NCI1 & NCI109 & Avg. \\
\midrule
 & $\times$ & $\times$ & 30.50$\pm$1.16 & 62.49$\pm$1.95 & 61.68$\pm$0.93 & 51.56 \\
GraphCL & \textbf{\checkmark} & $\times$ & 33.14$\pm$1.38 & 64.38$\pm$2.73 & 61.95$\pm$0.61 & 53.16 \\
 & $\times$ & \textbf{\checkmark} & \underline{35.81$\pm$2.80} & \underline{67.22$\pm$2.31} & \underline{66.22$\pm$0.78} & \underline{56.42} \\
 & \textbf{\checkmark} & \textbf{\checkmark} & \textbf{36.97$\pm$3.47} & \textbf{67.28$\pm$1.69} & \textbf{67.04$\pm$0.98} & \textbf{57.10} \\

 \midrule
 
 & $\times$ & $\times$ & 27.07$\pm$1.04 & 61.27$\pm$3.64 & 62.12$\pm$1.10 & 50.15 \\
SimGRACE & \textbf{\checkmark} & $\times$ & 27.44$\pm$0.95 & 61.49$\pm$3.68 & 61.93$\pm$1.22 & 50.29 \\
 & $\times$ & \textbf{\checkmark} & \underline{32.36$\pm$2.60} & \underline{66.41$\pm$1.55} & \underline{66.18$\pm$1.88} & \underline{54.98} \\
 & \textbf{\checkmark} & \textbf{\checkmark} & \textbf{34.86$\pm$3.00} & \textbf{67.10$\pm$1.45} & \textbf{67.23$\pm$1.72} & \textbf{56.40} \\

 \midrule
 
 & $\times$ & $\times$ & 29.08$\pm$1.35 & 56.85$\pm$4.35 & 62.27$\pm$0.78 & 49.40 \\
LP-GPPT & \textbf{\checkmark} & $\times$ & 31.28$\pm$1.83 & 59.08$\pm$3.22 & 63.17$\pm$0.98 & 51.18 \\
 & $\times$ & \textbf{\checkmark} & \underline{34.94$\pm$2.33} & \textbf{65.91$\pm$2.49} & \underline{65.66$\pm$1.14} & \underline{55.50} \\
 & \textbf{\checkmark} & \textbf{\checkmark} & \textbf{37.19$\pm$2.79} & \underline{65.36$\pm$2.33} & \textbf{66.17$\pm$1.96} & \textbf{56.24} \\

 \midrule
 
 & $\times$ & $\times$ & 31.33$\pm$3.22 & 62.09$\pm$2.31 & 60.19$\pm$1.71 & 51.20 \\
LP-GraphPrompt & \textbf{\checkmark} & $\times$ & 32.31$\pm$1.38 & 62.61$\pm$1.79 & 62.08$\pm$1.65 & 52.33 \\
 & $\times$ & \textbf{\checkmark} & \underline{36.27$\pm$2.68} & \underline{64.59$\pm$1.95} & \underline{63.64$\pm$1.98} & \underline{54.83} \\
 & \textbf{\checkmark} & \textbf{\checkmark} & \textbf{36.39$\pm$2.09} & \textbf{65.58$\pm$1.15} & \textbf{65.41$\pm$1.83} & \textbf{55.79} \\
\bottomrule
\end{tabular}
\end{adjustbox}
\end{table}

\paragraph{Effect of the Number of Prompt Bases.}
\label{subsubsec:num_prompt}

\begin{figure}[!ht]
\centering
\begin{subfigure}[b]{\linewidth}
\centering
\includegraphics[width=0.7\linewidth]{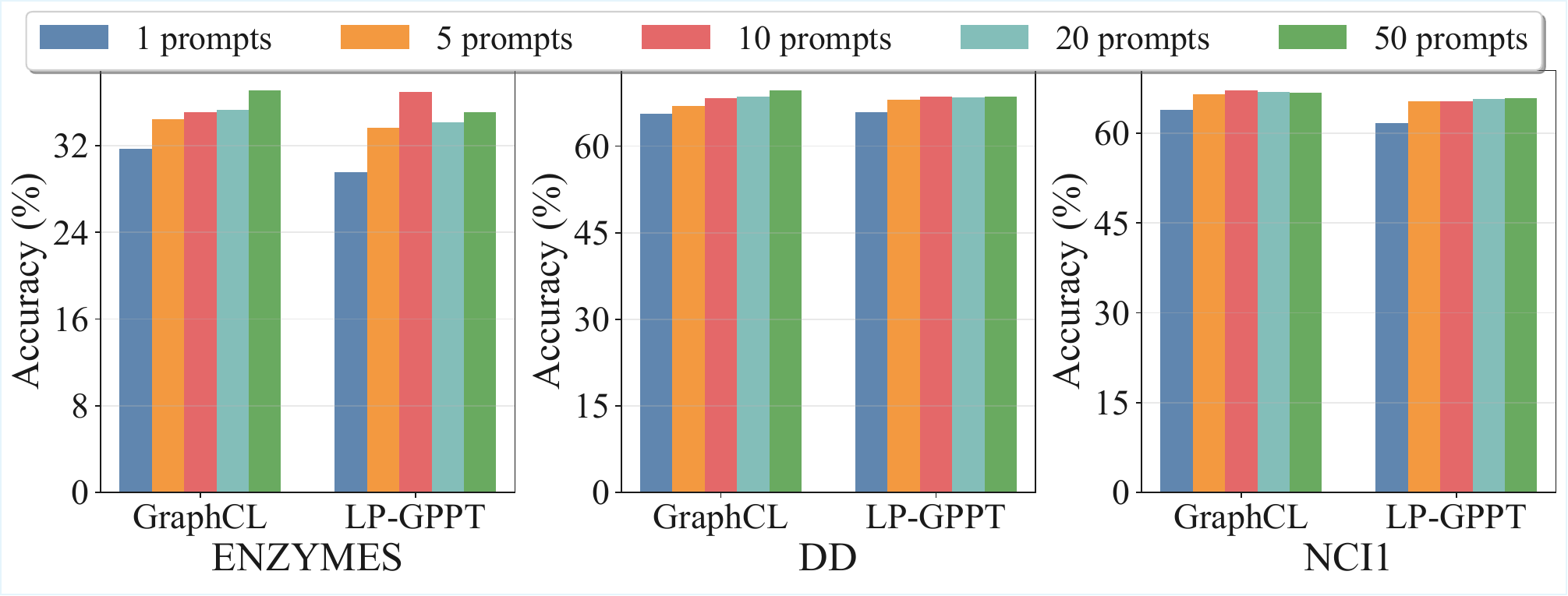}
\end{subfigure}
\caption{Performance of MAGPrompt+ across different numbers of prompt bases $M_l$ on ENZYMES, DD and NCI1 datasets.}
\label{fig:mix_num_prompt}
\end{figure}

We study the expressiveness–efficiency trade-off in MAGPrompt+ by varying the number of prompt bases $M_l$ per layer. As shown in Figure~\ref{fig:mix_num_prompt}, increasing $M_l$ yields consistent performance gains initially, followed by saturation, indicating that a modest number of prompt bases (typically 10-20) suffices to capture diverse interaction patterns. Compared to EdgePrompt, which benefits from a small number of prompts (typically 5–10), MAGPrompt+ can effectively leverage larger $M_l$ due to the prompt-collapse regularization $\mathcal{L}_{\mathrm{pc}}$, which promotes balanced prompt utilization and prevents degeneracy.

\begin{figure}[!ht]
\centering
\begin{subfigure}[b]{\linewidth}
\centering
\includegraphics[width=0.7\linewidth]{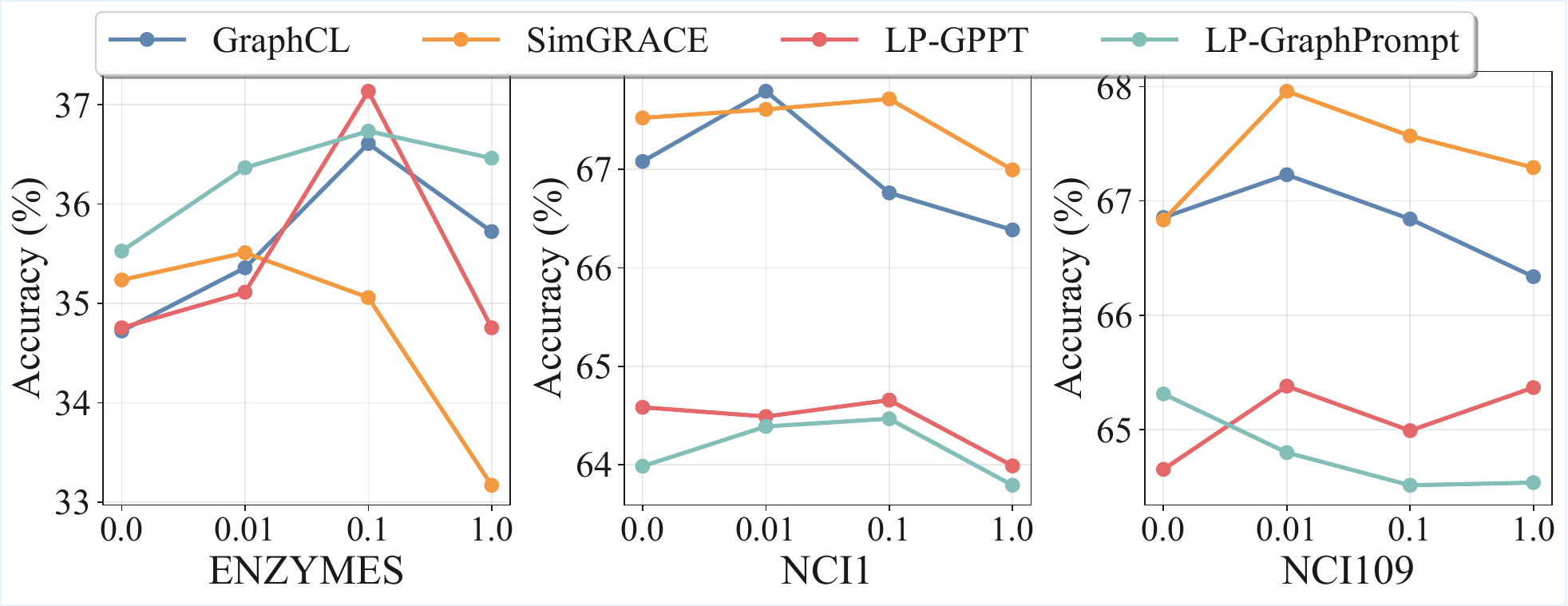}
\end{subfigure}
\caption{Performance of MAGPrompt+ across different $\lambda_{\mathrm{pc}}$ across ENZYMES, NCI1 and NCI109 datasets.}
\label{fig:graph_lambda_1}
\end{figure}

\paragraph{Effect of Prompt-collapse Regularization and Gate Parameter.}
\label{subsubsec:lambda_pc_beta}

\begin{table}[!ht]
\centering
\caption{Average classification Accuracy (\%) on 50-shot graph classification tasks of MAGPrompt+ over ENZYMES, DD and NCI1 datasets, across different $\beta$ selections. The best and second-best results are highlighted in bold and underlined, respectively.}
\label{tab:beta}
\begin{adjustbox}{width=0.55\textwidth}
\begin{tabular}{l|c|ccc|c}
\toprule
Pre-training & $\beta$ & ENZYMES & DD & NCI1 & Avg. \\
\midrule
 & 0.00 & 35.31$\pm$1.39 & \textbf{70.21$\pm$1.44} & 66.44$\pm$1.82 & 57.32 \\
 & 0.25 & 36.06$\pm$2.36 & 68.72$\pm$2.47 & 67.09$\pm$1.38 & 57.29 \\
GraphCL & 0.50 & \textbf{36.97$\pm$3.47} & \underline{69.40$\pm$2.56} & \underline{67.28$\pm$1.69} & \textbf{57.88} \\
 & 0.75 & \underline{36.87$\pm$2.35} & 69.21$\pm$3.42 & \textbf{67.30$\pm$2.22} & \underline{57.79} \\
 & 1.00 & 35.81$\pm$2.80 & 68.74$\pm$2.18 & 67.22$\pm$2.31 & 57.26 \\

\midrule

 & 0.00 & 34.36$\pm$2.59 & \textbf{69.51$\pm$1.44} & 64.26$\pm$1.97 & 56.04 \\
 & 0.25 & 34.69$\pm$2.14 & 69.09$\pm$1.51 & 65.19$\pm$1.93 & 56.32 \\
LP-GPPT & 0.50 & \textbf{37.19$\pm$2.79} & \underline{69.13$\pm$1.81} & 65.36$\pm$2.33 & \textbf{57.23} \\
 & 0.75 & \underline{35.03$\pm$3.25} & 68.92$\pm$1.75 & \underline{65.46$\pm$2.11} & 56.47 \\
 & 1.00 & 34.94$\pm$2.33 & 68.83$\pm$1.48 & \textbf{65.91$\pm$2.49} & \underline{56.56} \\

\bottomrule
\end{tabular}
\end{adjustbox}
\end{table}

Figure~\ref{fig:graph_lambda_1} analyzes the sensitivity of MAGPrompt+ to the prompt-collapse regularization weight $\lambda_{\mathrm{pc}}$ on TUDataset benchmarks under different pre-training strategies. Moderate regularization consistently achieves the best performance, with $\lambda_{\mathrm{pc}}\in[0.01,0.1]$ yielding optimal or near-optimal results. Without regularization ($\lambda_{\mathrm{pc}}=0$), performance degrades due to prompt collapse, while overly strong regularization ($\lambda_{\mathrm{pc}}=1.0$) restricts edge-specific adaptivity. The observed trends are stable across all pre-training schemes, indicating robust behavior.
Table~\ref{tab:beta} further examines the residual gate parameter $\beta$. Moderate values ($\beta\in[0.25,0.75]$) consistently provide strong performance, with $\beta=0.5$ offering the most stable results overall. While $\beta=0$ performs best on DD--likely due to its large, dense, and regular graph structure--datasets with smaller or noisier graphs (e.g., ENZYMES and NCI1) benefit from non-zero $\beta$, highlighting a data-dependent trade-off between expressiveness and robustness.

\paragraph{Effect of Attention Hidden Dimention $d_a$.}

\begin{figure}[!ht]
\centering
\begin{subfigure}[b]{\linewidth}
\centering
\includegraphics[width=0.65\linewidth]{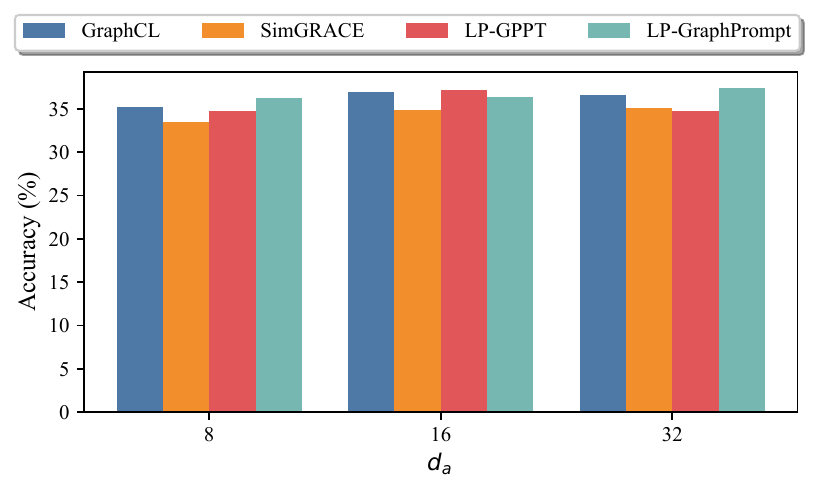}
\end{subfigure}
\caption{Performance of MAGPrompt+ across different $d_a$ on ENZYMES dataset.}
\label{fig:att_dim}
\end{figure}

Figure~\ref{fig:att_dim} analyzes the sensitivity of MAGPrompt+ to the attention head dimension $d_a$ on the ENZYMES dataset.
Across all pre-training strategies, performance is stable for $d_a \in \{8,16,32\}$, with the best or near-best results typically achieved at $d_a=16$.
This suggests that a moderate number of attention heads is sufficient, while increasing $d_a$ further yields diminishing returns. %  to capture effective message-adaptive reweighting

\section{Conclusion}

We proposed {MAGPrompt}, a message-adaptive graph prompt tuning framework for {parameter-efficient} adaptation of pre-trained GNNs.
In contrast to existing graph prompting methods that modify inputs, representations, or topology, our approach %directly
modulates per-neighbor message contributions during message passing via learnable, task-adaptive gates, while keeping the backbone frozen.
Extensive experiments on node- and graph-level benchmarks demonstrate consistent improvements over prior graph prompt tuning methods under few-shot and full-shot settings, with performance competitive with fine-tuning across diverse pre-training strategies.
These results highlight the effectiveness of adapting neighborhood message weighting as a principled direction for prompt-based graph learning.

%=============================================================================%

%Bibliography
\bibliographystyle{unsrt}  
\bibliography{references}

%%%%%%%%%%%%%%%%%%%%%%%%%%%%%%%%%%%%%%%%%%%%%%%%%%%%%%%%%%%%%%%%%%%%%%%%%%%%%%%
%%%%%%%%%%%%%%%%%%%%%%%%%%%%%%%%%%%%%%%%%%%%%%%%%%%%%%%%%%%%%%%%%%%%%%%%%%%%%%%
% APPENDIX
%%%%%%%%%%%%%%%%%%%%%%%%%%%%%%%%%%%%%%%%%%%%%%%%%%%%%%%%%%%%%%%%%%%%%%%%%%%%%%%
%%%%%%%%%%%%%%%%%%%%%%%%%%%%%%%%%%%%%%%%%%%%%%%%%%%%%%%%%%%%%%%%%%%%%%%%%%%%%%%
\newpage
\appendix
\onecolumn

\section{Proofs}

\subsection{Proof of Proposition 1}
\label{app:proof_1}

We prove that MAGPrompt+ (i) preserves the input graph structure and (ii) preserves the permutation
equivariance (node-level) and permutation invariance (graph-level) properties of the frozen backbone GNN.

\paragraph{Permutation operator and equivariance.}
Let $\pi$ be any permutation of the node indices $\{1,\ldots,n\}$, and let $\mathbf{P}\in\{0,1\}^{n\times n}$ be the
corresponding permutation matrix. For any learned node embedding matrix $\mathbf{H}\in\mathbb{R}^{n\times d}$,
$(\mathbf{P}\mathbf{H})_{\pi(i)}=\mathbf{H}_i$.
We represent the graph by a original node feature matrix $\mathbf{X}$ and an adjacency matrix $\mathbf{A}$ (and optionally edge attributes $\mathbf{E}$).
The permuted graph is given by
\begin{equation}
\mathbf{X}^{\pi}=\mathbf{P}\mathbf{X}, \qquad
\mathbf{A}^{\pi}=\mathbf{P}\mathbf{A}\mathbf{P}^{\top}, \qquad
\mathbf{E}^{\pi}=\mathbf{P}\mathbf{E}\mathbf{P}^{\top} \ \ (\text{if applicable}).
\end{equation}
A node encoder $F$ is \emph{permutation equivariant} if
\begin{equation}
F(\mathbf{A}^{\pi},\mathbf{X}^{\pi},\mathbf{E}^{\pi})=\mathbf{P}F(\mathbf{A},\mathbf{X},\mathbf{E}).
\label{eq:app_perm_equiv_def}
\end{equation}
A graph-level representation is \emph{permutation invariant} if a readout $\rho$ satisfies
$\rho(\mathbf{P}\mathbf{H})=\rho(\mathbf{H})$ for all $\mathbf{H}$.

\paragraph{Part (i): Structure preservation.}
MAGPrompt+ does not add, remove, or rewire edges. All computations are performed on the original edge set $\mathcal{E}$,
so the neighborhood $\mathcal{N}(v_i)$ is unchanged for every node $v_i$. Hence the underlying graph structure is preserved.

\paragraph{Part (ii): Node-level permutation equivariance.}
Consider a frozen message-passing GNN layer $l$ with the generic update
\begin{equation}
\mathbf{h}_i^{(l)}=
\textsc{UPD}^{(l)}\!\left(
\mathbf{h}_i^{(l-1)},
\textsc{AGG}^{(l)}\!\left(
\left\{
\mathbf{m}_{ij}^{(l)} \mid v_j\in \mathcal{N}(v_i)
\right\}
\right)
\right),
\label{eq:app_mp_update}
\end{equation}
where $\textsc{AGG}^{(l)}(\cdot)$ is a permutation-invariant multiset operator (e.g., sum/mean/max), and $\textsc{UPD}^{(l)}(\cdot)$ is
applied pointwise with shared parameters.
These are the standard sufficient conditions for permutation equivariance of message-passing GNNs.

MAGPrompt+ modifies messages by applying an edge-wise gate and injecting an additive prompt:
\begin{equation}
\tilde{\mathbf{m}}_{ij}^{(l)}=
a_{ij}^{(l)}\cdot \mathbf{m}_{ij}^{(l)} + \mathbf{p}_{ij}^{(l)},
\label{eq:app_magp_message}
\end{equation}
\begin{equation}
\mathbf{h}_i^{(l)}=
\textsc{UPD}^{(l)}\!\left(
\mathbf{h}_i^{(l-1)},
\textsc{AGG}^{(l)}\!\left(
\left\{
\tilde{\mathbf{m}}_{ij}^{(l)} \mid v_j\in \mathcal{N}(v_i)
\right\}
\right)
\right),
\label{eq:app_mp_update_appendix}
\end{equation}
where $\mathbf{m}_{ij}^{(l)}$ denotes the base message computed by the frozen backbone,
$a_{ij}^{(l)}\in\mathbb{R}$ is a scalar gate, and $\mathbf{p}_{ij}^{(l)}\in\mathbb{R}^{d_l}$ is the (possibly compositional)
edge prompt.

\paragraph{Equivariance of the gate and prompt.}
Under a permutation $\pi$, the edge $(i,j)$ maps to $(\pi(i),\pi(j))$ and the neighbor sets permute as
$\mathcal{N}^{\pi}(\pi(i))=\{\pi(j)\,:\, j\in\mathcal{N}(i)\}$.
By construction, MAGPrompt+ computes $a_{ij}^{(l)}$ and $\mathbf{p}_{ij}^{(l)}$ from (i) node representations using shared
linear maps and pointwise operations, and (ii) neighbor-wise softmax normalization.
All of these operations commute with reindexing of nodes:
\begin{equation}
a_{\pi(i)\pi(j)}^{(l),\pi}=a_{ij}^{(l)}, \qquad
\mathbf{p}_{\pi(i)\pi(j)}^{(l),\pi}=\mathbf{p}_{ij}^{(l)}.
\label{eq:app_gate_prompt_equiv}
\end{equation}
In addition, the frozen backbone message ${\mathbf{m}}_{ij}^{(l)}$ is permutation equivariant by the standard
message-passing assumption, hence
${\mathbf{m}}_{\pi(i)\pi(j)}^{(l),\pi}={\mathbf{m}}_{ij}^{(l)}$.
Combining with~\eqref{eq:app_magp_message} yields
\begin{equation}
\tilde{\mathbf{m}}_{\pi(i)\pi(j)}^{(l),\pi}=\tilde{\mathbf{m}}_{ij}^{(l)}.
\label{eq:app_message_equiv}
\end{equation}

\paragraph{Equivariance of aggregation and update.}
For each node $i$, the multiset of incoming messages
$\{\mathbf{m}_{ij}^{(l)}: j\in\mathcal{N}(i)\}$ is reindexed into
$\{\mathbf{m}_{\pi(i)\pi(j)}^{(l),\pi}: \pi(j)\in\mathcal{N}^{\pi}(\pi(i))\}$ under permutation.
Since $\textsc{AGG}^{(l)}(\cdot)$ is permutation-invariant over multisets, the aggregated message is preserved under reindexing.
Finally, because $\textsc{UPD}^{(l)}(\cdot)$ is applied pointwise with shared parameters, the resulting node representations satisfy
\begin{equation}
\mathbf{H}^{(l),\pi}=\mathbf{P}\mathbf{H}^{(l)}.
\end{equation}
Thus, MAGPrompt+ is permutation equivariant at the node level.

\paragraph{Graph-level permutation invariance.}
For graph-level tasks, MAGPrompt+ produces node embeddings $\mathbf{H}$ that are permutation equivariant.
Applying any permutation-invariant readout $\rho$ (e.g., sum/mean/max pooling) yields
$\rho(\mathbf{P}\mathbf{H})=\rho(\mathbf{H})$, so the graph representation is permutation invariant.

\paragraph{Conclusion.}
MAGPrompt+ preserves the graph structure and maintains the permutation equivariance/invariance properties of the underlying
message-passing GNN. \hfill$\square$

%-------------------------------------%

\subsection{Proof of Proposition 2}
\label{app:proof_2}

Fix a layer $l$ and a mini-batch of graphs $\mathcal{B}$ with edge set $\mathcal{E}_{\mathcal{B}}$.
For each edge $(i,j)\in\mathcal{E}_{\mathcal{B}}$, MAGPrompt+ produces mixture weights
$\boldsymbol{\pi}^{(l)}_{ij}\in\mathbb{R}^{M_l}$ such that $\pi^{(l)}_{ijm}\ge 0$ and
$\sum_{m=1}^{M_l}\pi^{(l)}_{ijm}=1$.
Define the aggregated usage of the $m$-th prompt basis vector as
\begin{equation}
s_m^{(l)} \;=\; \sum_{(i,j)\in\mathcal{E}_{\mathcal{B}}}\pi_{ijm}^{(l)},\qquad
\mathbf{s}^{(l)} = [s_1^{(l)},\dots,s_{M_l}^{(l)}]^\top .
\end{equation}
Summing over $m$ and using $\sum_{m}\pi^{(l)}_{ijm}=1$ gives
\begin{equation}
\sum_{m=1}^{M_l} s_m^{(l)}
= \sum_{(i,j)\in\mathcal{E}_{\mathcal{B}}}\sum_{m=1}^{M_l}\pi_{ijm}^{(l)}
= |\mathcal{E}_{\mathcal{B}}|.
\end{equation}
Hence the mean usage
\begin{equation}
\bar{s}^{(l)} \;=\; \frac{1}{M_l}\sum_{m=1}^{M_l}s_m^{(l)} \;=\; \frac{|\mathcal{E}_{\mathcal{B}}|}{M_l}
\end{equation}
is fixed for the given batch.

MAGPrompt+ defines the prompt-collapse regularizer
\begin{equation}
\mathcal{L}_{\mathrm{pc}}^{(l)}
=\frac{1}{M_l}\sum_{m=1}^{M_l}\frac{\bigl(s_m^{(l)}-\bar{s}^{(l)}\bigr)^2}{\bigl(\bar{s}^{(l)}\bigr)^2+\epsilon},
\qquad \epsilon>0 .
\end{equation}
Since $\bigl(\bar{s}^{(l)}\bigr)^2+\epsilon>0$ is constant w.r.t.\ $\mathbf{s}^{(l)}$, minimizing
$\mathcal{L}_{\mathrm{pc}}^{(l)}$ is equivalent to minimizing
$\sum_{m=1}^{M_l}\bigl(s_m^{(l)}-\bar{s}^{(l)}\bigr)^2=\|\mathbf{s}^{(l)}-\bar{s}^{(l)}\mathbf{1}\|_2^2$.
This quantity is nonnegative and equals $0$ if and only if
\begin{equation}
s_1^{(l)}=\cdots=s_{M_l}^{(l)}=\bar{s}^{(l)},
\end{equation}
i.e., all prompt basis vectors are used equally often in aggregate across edges in the batch.
Therefore, $\mathcal{L}_{\mathrm{pc}}^{(l)}$ penalizes imbalanced prompt utilization and discourages degenerate
solutions in which a small subset of prompts dominates globally.

Finally, note that $\mathcal{L}_{\mathrm{pc}}^{(l)}$ depends only on the aggregated counts $\mathbf{s}^{(l)}$, not on any
single $\boldsymbol{\pi}^{(l)}_{ij}$. Thus it encourages balanced \emph{global} usage without forcing
$\boldsymbol{\pi}^{(l)}_{ij}$ to be uniform on every edge, preserving edge-wise adaptivity. \hfill$\square$

%-------------------------------------%

\subsection{Complexity Analysis}
\label{app:complexity}

We analyze the computational complexity of MAGPrompt and MAGPrompt+ on common message-passing GNN backbones (e.g., GCN, GIN, and GINE). Let $\mathcal{G}=(\mathcal{V},\mathcal{E})$ be a graph with $n=|\mathcal{V}|$ nodes and $m=|\mathcal{E}|$ edges, and let $d$ denote the hidden dimension. We assume sparse adjacency and standard neighborhood aggregation.

\paragraph{Backbone GNN (frozen encoder).}
A typical message-passing layer consists of (i) edge-wise message computation and aggregation and (ii) a node-wise transformation (linear layer or MLP). This yields the per-layer time complexity $\mathcal{O}(m\,d \;+\; n\,d^2),$
% \begin{equation}
% T_{\text{backbone}} = \mathcal{O}(m\,d) \;+\; \mathcal{O}(n\,d^2),
% \end{equation}
where the $\mathcal{O}(m\,d)$ term covers sparse edge operations and aggregation, and the $\mathcal{O}(n\,d^2)$ term accounts for dense node-wise transformations (e.g., MLPs in GCN/GIN/GINE layers).

\paragraph{MAGPrompt.}
\label{app:magprompt_complex}
MAGPrompt adds (i) a node projection in Eq.~\eqref{eq:proj_b} with cost $\mathcal{O}(n\,d\,d_a)$ and
(ii) head-wise attention scoring and neighborhood-wise softmax in Eqs.~\eqref{eq:node_prj_att} -- \eqref{eq:gate_beta} with cost $\mathcal{O}(n\,d_a + m\,d_a)$ per layer.
Prompted message scaling and injection in Eq.~\eqref{eq:prompted_msg_mag} is element-wise and costs $\mathcal{O}(m\,d)$, matching the backbone edge term.
When $d_a \ll d$ and treated as a small constant, the overall asymptotic complexity remains
$\mathcal{O}(m\,d + n\,d^2)$, identical to that of the frozen backbone GNN.

\paragraph{MAGPrompt+.}
\label{app:magprompt_plus_complex}
MAGPrompt+ retains the same gating cost as MAGPrompt and additionally computes an edge-adaptive prompt by mixing $M_l$ basis vectors.
For each edge, mixture computation Eq.~\eqref{eq:mix_logits} and prompt composition Eq.~\eqref{eq:edge_prompt_comp} incur $\mathcal{O}(M_l(d+d_a))$ cost, yielding an additional $\mathcal{O}(m\,M_l(d+d_a))$ term per layer.
Since $M_l$ is chosen as a small constant in practice, MAGPrompt+ remains linear in the number of edges and preserves the scalability of the backbone.

\paragraph{Parameter overhead.}
Let $L$ be the number of layers.
MAGPrompt introduces one prompt vector $p^{(l)}\in\mathbb{R}^{d}$ per layer and lightweight gating parameters of size $\mathcal{O}(d\,d_a)$, resulting in $\mathcal{O}(L(d + d\,d_a))$ trainable parameters.
MAGPrompt+ additionally maintains $M_l$ prompt vectors per layer and a small projection for prompt selection, contributing $\mathcal{O}(L\,M_l(d+d_a))$ parameters.
Under the practical setting $d_a, M_l \ll d$, both variants are substantially more parameter-efficient than fine-tuning the full backbone.

%-------------------------------------%

\section{More Experimental Setup Descriptions}
\subsection{Datasets}
\label{app:data}

We evaluate our methods on both node- and graph-level benchmarks.
For node classification (Table~\ref{tab:node_dataset_stats}), we use Cora, CiteSeer, Pubmed~\cite{yang2016revisiting}, ogbn-arxiv~\cite{hu2020open}, and Flickr~\cite{graphsainticlr20} under 5-shot classification settings, following EdgePrompt setting~\cite{fu2025edge}.
For graph classification, we adopt five TUDataset~\cite{morris2020tudataset} benchmarks (Table~\ref{tab:tu_dataset_stats}), including ENZYMES, DD, NCI1, NCI109, and Mutagenicity under 50-shot classification settings as decribed in EdgePrompt~\cite{fu2025edge}.
In addition, we evaluate on four MoleculeNet~\cite{wu2018moleculenet} datasets (Table~\ref{tab:moleculenet_stats}), which contain BACE, BBBP, SIDER, and ClinTox in the standard full-shot setting, following the configuration of GPF~\cite{fang2023universal}; the data curation and scaffold split of~\cite{nguyen2024smiles}.
MoleculeNet datasets include edge attributes, while the remaining benchmarks contain only node features and graph connectivity.

% ------------------------------------------------------------
% Table 1: Node-level datasets
% ------------------------------------------------------------
\begin{table}[!ht]
\centering
\caption{Details of node-level datasets.}
\label{tab:node_dataset_stats}
\begin{adjustbox}{width=0.65\linewidth}
\begin{tabular}{lcccccc}
\toprule
\textbf{Dataset} & \#Graphs & \#Nodes & \#Edges & \#Features & \#Classes & Task Level \\
\midrule
Cora        & 1 & 2,708   & 10,556    & 1,433 & 7  & Node \\
CiteSeer    & 1 & 3,327   & 9,104     & 3,703 & 6  & Node \\
Pubmed      & 1 & 19,717  & 88,648    & 500   & 3  & Node \\
Flickr      & 1 & 89,250  & 899,756   & 500   & 7  & Node \\
ogbn-arxiv  & 1 & 169,343 & 1,166,243 & 128   & 40 & Node \\
\bottomrule
\end{tabular}
\end{adjustbox}
\end{table}

% ------------------------------------------------------------
% Table 2: TUDataset graph-level datasets
% ------------------------------------------------------------
\begin{table}[!ht]
\centering
\caption{Details of TUDataset graph-level benchmarks.}
\label{tab:tu_dataset_stats}
\begin{adjustbox}{width=0.7\linewidth}
\begin{tabular}{lcccccc}
\toprule
\textbf{Dataset} & \#Graphs & Avg. Nodes & Avg. Edges & \#Features & \#Classes & Task Level \\
\midrule
ENZYMES        & 600   & 32.63  & 124.27   & 3  & 6 & Graph \\
DD             & 1,178 & 284.32 & 1,431.32 & 89 & 2 & Graph \\
NCI1           & 4,110 & 29.87  & 64.60    & 37 & 2 & Graph \\
NCI109         & 4,127 & 29.68  & 64.26    & 38 & 2 & Graph \\
Mutagenicity   & 4,337 & 30.32  & 61.54    & 14 & 2 & Graph \\
\bottomrule
\end{tabular}
\end{adjustbox}
\end{table}

% ------------------------------------------------------------
% Table 3: MoleculeNet datasets
% ------------------------------------------------------------
\begin{table}[!ht]
\centering
\caption{Details of MoleculeNet datasets.}
\label{tab:moleculenet_stats}
\begin{adjustbox}{width=0.7\linewidth}
\begin{tabular}{lccccc}
\toprule
\textbf{Dataset} & \#Tasks & Task type & \#Original samples & \#Refined samples \\
\midrule
% ESOL           & 1  & Regression              & 1128 & 1115 \\
% FreeSolv       & 1  & Regression              & 642  & 635  \\
% Lipophilicity  & 1  & Regression              & 4200 & 4100 \\
% \midrule
BACE           & 1  & Binary classification    & 1513 & 1454 \\
BBBP           & 1  & Binary classification    & 2050 & 1760 \\
\midrule
ClinTox        & 2  & Multi-label classification & 1484 & 1349 \\
SIDER          & 27 & Multi-label classification & 1427 & 1225 \\
% Tox21          & 12 & Multi-label classification & 7831 & 7381 \\
\bottomrule
\end{tabular}
\end{adjustbox}
\end{table}

\subsection{Pre-training strategies}
\label{app:pretrain}

We consider eight representative self-supervised graph pre-training methods spanning both contrastive and predictive paradigms, following the pre-training practice of EdgePrompt~\cite{fu2025edge} and GPF~\cite{fang2023universal}:

\begin{itemize}
    \item \textbf{DGI}~\cite{veličković2018deep} maximizes mutual information between local node representations and a global graph summary to capture global structural dependencies.

    \item \textbf{GraphCL}~\cite{you2020graph} is a contrastive framework that learns invariant graph representations by maximizing agreement between multiple augmented views of the same graph.

    \item \textbf{SimGRACE}~\cite{xia2022simgrace} generates correlated views by perturbing the GNN encoder parameters with Gaussian noise, avoiding explicit graph augmentations and improving robustness on sensitive domains.

    \item \textbf{AttrMasking}~\cite{Hu*2020Strategies} trains the model to predict masked node or edge attributes from their local neighborhoods.

    \item \textbf{ContextPred}~\cite{Hu*2020Strategies} learns structural semantics by predicting whether a subgraph matches its surrounding context.

    \item \textbf{EdgePred}~\cite{kipf2017semisupervised} reconstructs masked edges to encode intrinsic connectivity and proximity information.

    \item \textbf{LP-GPPT}~\cite{sun2022gppt} reformulates downstream node classification as a link prediction task using task and structure tokens to bridge the pre-training and adaptation objectives.

    \item \textbf{LP-GraphPrompt}~\cite{liu2023graphprompt} unifies pre-training and downstream learning through a subgraph similarity template, leveraging link prediction and learnable prompts at the readout stage.
\end{itemize}

%-------------------------------------%
\section{More Experimental Results}
\subsection{Node Classification \& Graph Classification}

% We provide more detailed experiments on both node- and graph-level classification tasks in Table~\ref{tab:node_class_full}, Table~\ref{tab:graph_class_full} and Table~\ref{tab:molecule_full}. They are detailed and provide a clearer picture of our methods compared to previous methods like GPF/GPF-plus, GPPT, GraphPrompt, ALL-in-One, EdgePrompt/EdgePrompt+, and GraphTop. These results highlight that our proposed MAGPrompt methods consistently outperform other methods on 5-shot node classification with a margin of more than 2\%, and 50-shot graph classification with a margin of more than 2\%. In a full-shot setting of MoleculeNet datasets, these results show that MAGPrompt outperform EdgePrompt and can be better than GPF and full fine-tuning in most of the datasets, indicating the effectiveness of our proposed methods.

We report the complete experimental results for both node-level and graph-level classification tasks in Tables~\ref{tab:node_class_full}, \ref{tab:graph_class_full}, and \ref{tab:molecule_full}.
These tables provide a comprehensive comparison between our proposed methods and representative graph prompt tuning baselines, including GPF/GPF-plus, GPPT, GraphPrompt, ALL-in-One, EdgePrompt/EdgePrompt+, and GraphTOP. 
%The original GraphTOP study was restricted to the Cora, PubMed, and Flickr benchmarks (Table~\ref{tab:node_class_full}). 
% , we extended the evaluation to CiteSeer and ogbn-arxiv using the authors' publicly available implementation\footnote{\href{https://github.com/xbfu/GraphTOP}{https://github.com/xbfu/GraphTOP}}
% . 
% However, our results on these additional datasets yield accuracies comparable to random initialization. This discrepancy is likely attributable to sensitivities within the edge rewiring mechanism when applied to these specific graph topologies. 
Across all 5-shot node classification benchmarks, MAGPrompt and MAGPrompt+ consistently achieve superior performance, yielding improvements on average compared to prior methods.
Similarly, in the 50-shot graph classification setting, our methods outperform existing prompt-based approaches.
On full-shot MoleculeNet benchmarks, MAGPrompt+ consistently surpasses EdgePrompt and achieves performance comparable to or exceeding GPF and full fine-tuning on most datasets, demonstrating the effectiveness and robustness of message-adaptive prompting across diverse pre-training strategies and task settings.

\begin{table*}
\centering
\caption{Average accuracy on 5-shot node classification tasks over five datasets. Best results are in bold; second-best are underlined.}
\label{tab:node_class_full}
\begin{adjustbox}{width=1\textwidth}
\begin{tabular}{l|c|ccccc|c}\toprule
Pre-training & Tuning & Cora & CiteSeer & Pubmed & ogbn-arxiv & Flickr & Avg. \\
\midrule
 & Linear Probe & 53.05$\pm$4.76 & 38.62$\pm$3.43 & 64.28$\pm$4.51 & 21.15$\pm$1.64 & 24.32$\pm$2.93 & 40.28 \\
 & GPPT & 50.96$\pm$6.67 & 39.50$\pm$1.67 & 60.47$\pm$4.75 & 17.99$\pm$1.14 & 24.35$\pm$1.84 & 38.65 \\
 & GraphPrompt & 55.71$\pm$4.62 & 40.81$\pm$2.11 & 63.47$\pm$2.23 & 21.03$\pm$1.92 & \textbf{26.08$\pm$3.44} & 41.42 \\
 & ALL-in-one & 38.00$\pm$4.17 & 40.27$\pm$2.09 & 58.61$\pm$3.49 & 16.42$\pm$2.98 & 25.08$\pm$3.44 & 35.68 \\
 & GPF & 58.52$\pm$4.07 & 43.55$\pm$2.80 & \underline{67.67$\pm$3.14} & 21.73$\pm$1.75 & 23.98$\pm$1.71 & 43.09 \\
GraphCL & GPF-plus & 52.24$\pm$4.59 & 38.47$\pm$3.27 & 64.30$\pm$4.58 & 21.03$\pm$1.96 & 25.32$\pm$2.02 & 40.27 \\
 & EdgePrompt & 58.60$\pm$4.46 & 43.31$\pm$3.23 & \textbf{67.76$\pm$3.01} & 21.90$\pm$1.71 & 24.83$\pm$2.78 & 43.28 \\
 & EdgePrompt+ & \underline{62.88$\pm$6.43} & \underline{46.20$\pm$0.99} & 67.41$\pm$5.25 & \underline{23.18$\pm$1.26} & 25.57$\pm$3.04 & \underline{45.05} \\
 & GraphTOP & 48.35$\pm$4.35 & 40.98$\pm$3.93 & 63.60$\pm$2.60 & 23.14$\pm$2.29 & 25.48$\pm$2.36 & 40.31 \\
 & MAGPrompt & 59.10$\pm$4.81 & 42.87$\pm$3.27 & 64.54$\pm$4.81 & 22.21$\pm$2.52 & 22.96$\pm$0.90 & 42.34 \\
 & MAGPrompt+ & \textbf{66.53$\pm$8.08} & \textbf{50.93$\pm$4.24} & 67.22$\pm$4.35 & \textbf{23.74$\pm$1.72} & \underline{25.85$\pm$3.29} & \textbf{46.85} \\

 \midrule
 
 & Linear Probe & 52.27$\pm$2.74 & 40.45$\pm$3.55 & 56.72$\pm$3.80 & 20.75$\pm$2.92 & 25.53$\pm$3.98 & 39.14 \\
 & GPPT & 52.07$\pm$7.65 & 40.25$\pm$3.29 & 58.65$\pm$5.12 & 17.76$\pm$1.80 & 23.37$\pm$4.66 & 38.42 \\
 & GraphPrompt & 51.42$\pm$2.80 & 41.74$\pm$2.22 & 55.98$\pm$2.94 & 20.48$\pm$2.57 & 25.88$\pm$3.81 & 39.10 \\
 & ALL-in-one & 34.64$\pm$4.06 & 38.95$\pm$2.35 & 54.18$\pm$4.70 & 16.72$\pm$2.90 & 27.68$\pm$4.58 & 34.43 \\
 & GPF & 58.23$\pm$4.19 & 44.87$\pm$4.35 & 61.55$\pm$3.79 & 21.86$\pm$2.91 & 26.51$\pm$4.69 & 42.60 \\
SimGRACE & GPF-plus & 52.27$\pm$3.34 & 41.02$\pm$3.49 & 56.95$\pm$3.86 & 21.44$\pm$3.77 & 28.35$\pm$5.50 & 40.01 \\
 & EdgePrompt & 58.37$\pm$4.51 & 43.94$\pm$4.15 & 61.10$\pm$3.69 & 21.85$\pm$2.54 & \textbf{30.12$\pm$5.04} & 43.08 \\
 & EdgePrompt+ & \underline{62.40$\pm$7.97} & \underline{46.62$\pm$2.53} & \underline{64.91$\pm$5.58} & \underline{22.74$\pm$2.34} & \underline{28.50$\pm$4.08} & \underline{45.03} \\
 & GraphTOP & 49.06$\pm$3.64 & 39.02$\pm$7.97 & 55.33$\pm$3.71 & 22.50$\pm$2.06 & 26.12$\pm$3.12 & 38.41 \\
 & MAGPrompt & 57.28$\pm$3.86 & 45.88$\pm$5.21 & 59.68$\pm$5.58 & 21.50$\pm$2.55 & 25.10$\pm$3.40 & 41.89 \\
 & MAGPrompt+ & \textbf{65.45$\pm$8.34} & \textbf{51.98$\pm$3.40} & \textbf{66.60$\pm$7.60} & \textbf{23.41$\pm$2.62} & 26.57$\pm$4.89 & \textbf{46.80} \\

 \midrule
 
 & Linear Probe & 28.65$\pm$4.82 & 26.77$\pm$2.03 & 40.14$\pm$5.69 & 11.57$\pm$1.91 & 28.39$\pm$7.44 & 27.10 \\
 & GPPT & 41.28$\pm$6.92 & 35.32$\pm$1.27 & 53.41$\pm$3.99 & 13.73$\pm$1.16 & 29.83$\pm$3.73 & 34.71 \\
 & GraphPrompt & 31.65$\pm$3.33 & 26.98$\pm$1.24 & 44.18$\pm$5.57 & 11.31$\pm$1.89 & 26.02$\pm$1.16 & 28.03 \\
 & ALL-in-one & 31.57$\pm$2.16 & 28.87$\pm$2.57 & 46.02$\pm$4.23 & 15.94$\pm$0.75 & 31.89$\pm$1.14 & 30.86 \\
 & GPF & 37.56$\pm$3.81 & 29.74$\pm$1.73 & 48.86$\pm$7.32 & 16.95$\pm$1.58 & 29.68$\pm$6.73 & 32.56 \\
LP-GPPT & GPF-plus & 28.87$\pm$3.18 & 26.65$\pm$1.91 & 40.32$\pm$5.77 & 11.78$\pm$1.55 & 29.41$\pm$6.79 & 27.41 \\
 & EdgePrompt & 37.26$\pm$4.53 & 29.83$\pm$1.01 & 47.20$\pm$7.06 & 17.22$\pm$1.31 & 31.17$\pm$6.58 & 32.54 \\
 & EdgePrompt+ & \underline{56.41$\pm$3.62} & \underline{43.49$\pm$2.62} & \underline{61.51$\pm$4.91} & \underline{17.78$\pm$2.16} & \underline{32.70$\pm$6.21} & \underline{42.38} \\
 & GraphTOP & 28.21$\pm$2.76 & 25.04$\pm$1.81 & 40.20$\pm$3.04 & 17.02$\pm$3.04 & 31.12$\pm$4.12 & 28.32 \\
 & MAGPrompt & 36.07$\pm$3.93 & 29.96$\pm$2.75 & 44.89$\pm$2.23 & \textbf{18.29$\pm$2.17} & 32.69$\pm$5.12 & 32.38 \\
 & MAGPrompt+ & \textbf{64.09$\pm$3.98} & \textbf{51.71$\pm$3.50} & \textbf{62.58$\pm$7.70} & 17.66$\pm$2.10 & \textbf{33.29$\pm$5.36} & \textbf{45.87} \\

 \midrule
 
 & Linear Probe & 59.00$\pm$5.74 & 44.54$\pm$4.44 & 72.09$\pm$5.70 & 31.28$\pm$1.50 & 27.83$\pm$4.77 & 46.95 \\
 & GPPT & 54.29$\pm$7.90 & 45.81$\pm$3.54 & 66.56$\pm$4.06 & 25.34$\pm$1.85 & 28.41$\pm$3.68 & 44.08 \\
 & GraphPrompt & 60.22$\pm$4.04 & 47.07$\pm$3.09 & 73.13$\pm$5.07 & 32.40$\pm$1.30 & 28.10$\pm$3.27 & 48.18 \\
 & ALL-in-one & 42.55$\pm$2.99 & 44.36$\pm$2.52 & 67.66$\pm$6.38 & 15.22$\pm$3.04 & 31.79$\pm$6.19 & 40.32 \\
 & GPF & 62.62$\pm$6.40 & 49.02$\pm$4.53 & \underline{73.62$\pm$6.42} & 31.88$\pm$1.08 & 28.98$\pm$5.30 & 49.22 \\
LP-GraphPrompt & GPF-plus & 58.23$\pm$5.68 & 44.60$\pm$4.47 & 72.15$\pm$5.64 & 31.58$\pm$1.09 & 28.96$\pm$4.63 & 47.10 \\
 & EdgePrompt & 62.74$\pm$6.77 & 48.69$\pm$4.36 & 73.60$\pm$5.14 & \underline{32.67$\pm$1.83} & 29.81$\pm$3.59 & 49.50 \\
 & EdgePrompt+ & \underline{64.47$\pm$7.04} & \underline{49.71$\pm$2.25} & \textbf{73.72$\pm$5.10} & 31.41$\pm$1.88 & \underline{32.09$\pm$4.93} & \textbf{50.28} \\
 & GraphTOP & 51.76$\pm$4.97 & 41.13$\pm$1.59 & 70.65$\pm$4.56 & \textbf{32.76$\pm$2.41} & 27.94$\pm$3.21 & 44.85 \\
 & MAGPrompt & 63.07$\pm$5.73 & 47.70$\pm$2.96 & 67.68$\pm$7.05 & 31.75$\pm$1.75 & 29.38$\pm$5.06 & 47.92 \\
 & MAGPrompt+ & \textbf{65.40$\pm$7.43} & \textbf{52.82$\pm$4.18} & 67.73$\pm$5.69 & 31.73$\pm$2.17 & \textbf{32.55$\pm$5.77} & \underline{50.05} \\
\bottomrule
\end{tabular}
\end{adjustbox}
\end{table*}

\begin{table*}
\centering
\caption{Average accuracy on 50-shot graph classification tasks over TUDataset. Best results are in bold; second-best are underlined.}
\label{tab:graph_class_full}
\begin{adjustbox}{width=1\textwidth}
\begin{tabular}{l|c|ccccc|c}
\toprule
Pre-training & Tuning & ENZYMES & DD & NCI1 & NCI109 & Mutagenicity & Avg. \\
\midrule
 & Linear Probe & 30.50$\pm$1.16 & 62.89$\pm$2.19 & 62.49$\pm$1.95 & 61.68$\pm$0.93 & 66.62$\pm$1.87 & 56.84 \\
 & GraphPrompt & 27.83$\pm$1.61 & 64.33$\pm$1.79 & 63.19$\pm$1.71 & 62.18$\pm$0.48 & \underline{67.62$\pm$0.65} & 57.03 \\
 & ALL-in-one & 25.92$\pm$0.55 & 66.54$\pm$1.82 & 57.52$\pm$2.61 & 62.74$\pm$0.78 & 63.43$\pm$2.53 & 55.23 \\
 & GPF & 30.08$\pm$1.25 & 64.54$\pm$2.22 & 62.66$\pm$1.83 & 62.29$\pm$0.90 & 66.54$\pm$1.85 & 57.22 \\
GraphCL & GPF-plus & 31.00$\pm$1.50 & 67.26$\pm$2.29 & 64.56$\pm$1.10 & 62.84$\pm$0.22 & 66.82$\pm$1.63 & 58.50 \\
 & EdgePrompt & 29.50$\pm$1.57 & 64.16$\pm$2.13 & 63.05$\pm$2.11 & 62.59$\pm$0.93 & 66.87$\pm$1.88 & 57.23 \\
 & EdgePrompt+ & \underline{34.00$\pm$1.25} & \underline{67.98$\pm$2.05} & \underline{66.30$\pm$2.54} & \underline{66.52$\pm$0.91} & 67.47$\pm$2.37 & \underline{60.45} \\
 & MAGPrompt & 31.72$\pm$2.53 & 65.49$\pm$2.64 & 63.62$\pm$2.69 & 62.96$\pm$2.09 & 67.56$\pm$1.98 & 58.27 \\
 & MAGPrompt+ & \textbf{36.97$\pm$3.47} & \textbf{69.40$\pm$2.56} & \textbf{67.28$\pm$1.69} & \textbf{67.04$\pm$0.98} & \textbf{68.64$\pm$1.45} & \textbf{61.87} \\

 \midrule
 
 & Linear Probe & 27.07$\pm$1.04 & 61.77$\pm$2.40 & 61.27$\pm$3.64 & 62.12$\pm$1.10 & 67.36$\pm$0.71 & 55.92 \\
 & GraphPrompt & 26.87$\pm$1.47 & 62.58$\pm$1.84 & 62.45$\pm$1.52 & 62.41$\pm$0.69 & 68.03$\pm$0.78 & 56.47 \\
 & ALL-in-one & 25.73$\pm$1.18 & 65.16$\pm$1.47 & 58.52$\pm$1.59 & 62.01$\pm$0.66 & 64.43$\pm$1.00 & 55.17 \\
 & GPF & 28.53$\pm$1.76 & 65.64$\pm$0.70 & 61.45$\pm$3.13 & 61.90$\pm$1.26 & 67.19$\pm$0.74 & 56.94 \\
SimGRACE & GPF-plus & 27.33$\pm$2.01 & 67.20$\pm$1.56 & 61.61$\pm$2.89 & 62.84$\pm$0.23 & 67.69$\pm$0.64 & 57.33 \\
 & EdgePrompt & 29.33$\pm$2.30 & 63.97$\pm$2.14 & 62.02$\pm$3.02 & 62.02$\pm$1.03 & 67.55$\pm$0.85 & 56.98 \\
 & EdgePrompt+ & \underline{32.67$\pm$2.53} & \underline{67.72$\pm$1.62} & \underline{67.07$\pm$1.96} & \underline{66.53$\pm$1.30} & \underline{68.31$\pm$1.36} & \underline{60.46} \\
 & MAGPrompt & 28.94$\pm$3.84 & 66.83$\pm$2.21 & 61.62$\pm$3.38 & 62.70$\pm$1.30 & 67.75$\pm$0.93 & 57.57 \\
 & MAGPrompt+ & \textbf{34.86$\pm$3.00} & \textbf{68.66$\pm$1.63} & \textbf{67.10$\pm$1.45} & \textbf{67.23$\pm$1.72} & \textbf{69.60$\pm$1.33} & \textbf{61.49} \\

 \midrule
 
 & Linear Probe & 29.08$\pm$1.35 & 62.12$\pm$2.82 & 56.85$\pm$4.35 & 62.27$\pm$0.78 & 66.30$\pm$1.78 & 55.32 \\
 & GraphPrompt & 26.67$\pm$1.60 & 61.61$\pm$1.91 & 58.77$\pm$0.97 & 62.16$\pm$0.89 & 66.37$\pm$1.17 & 55.12 \\
 & ALL-in-one & 24.92$\pm$1.33 & 63.61$\pm$2.12 & 59.14$\pm$2.12 & 59.70$\pm$1.37 & 64.86$\pm$1.60 & 54.45 \\
 & GPF & 28.33$\pm$1.73 & 63.48$\pm$2.08 & 58.14$\pm$4.16 & 62.52$\pm$1.39 & 66.10$\pm$0.96 & 55.71 \\
LP-GPPT & GPF-plus & 29.25$\pm$1.30 & \underline{66.92$\pm$2.34} & 62.93$\pm$3.23 & 64.13$\pm$1.42 & 67.57$\pm$1.45 & 58.16 \\
 & EdgePrompt & 28.33$\pm$3.41 & 64.03$\pm$2.26 & 59.85$\pm$3.15 & 62.98$\pm$1.44 & 66.36$\pm$1.22 & 56.31 \\
 & EdgePrompt+ & \underline{32.75$\pm$2.26} & 66.16$\pm$1.60 & \underline{63.58$\pm$2.07} & \underline{65.15$\pm$1.60} & \underline{68.35$\pm$1.57} & \underline{59.20} \\
 & MAGPrompt & 29.61$\pm$3.13 & 65.95$\pm$1.81 & 61.48$\pm$3.21 & 63.70$\pm$1.67 & 67.57$\pm$1.25 & 57.66 \\
 & MAGPrompt+ & \textbf{37.19$\pm$2.79} & \textbf{69.13$\pm$1.81} & \textbf{65.36$\pm$2.33} & \textbf{66.17$\pm$1.96} & \textbf{70.29$\pm$0.56} & \textbf{61.63} \\

 \midrule
 
 & Linear Probe & 31.33$\pm$3.22 & 62.58$\pm$2.40 & 62.09$\pm$2.31 & 60.19$\pm$1.71 & 65.13$\pm$0.81 & 56.26 \\
 & GraphPrompt & 30.20$\pm$1.93 & 64.72$\pm$1.98 & 62.57$\pm$1.45 & 62.32$\pm$0.95 & 65.85$\pm$0.65 & 57.13 \\
 & ALL-in-one & 29.07$\pm$1.16 & 65.60$\pm$2.38 & 58.67$\pm$2.42 & 57.69$\pm$1.08 & 64.66$\pm$0.76 & 55.14 \\
 & GPF & 30.93$\pm$1.76 & 66.21$\pm$1.66 & 61.80$\pm$2.78 & 62.27$\pm$1.18 & 65.61$\pm$0.59 & 57.36 \\
LP-GraphPrompt & GPF-plus & 30.67$\pm$3.06 & 67.50$\pm$2.45 & 62.59$\pm$2.09 & 61.98$\pm$1.60 & 65.51$\pm$1.10 & 57.65 \\
 & EdgePrompt & 30.80$\pm$2.09 & 65.87$\pm$1.35 & 61.75$\pm$2.49 & 62.33$\pm$1.65 & 65.77$\pm$0.90 & 57.30 \\
 & EdgePrompt+ & \underline{33.27$\pm$2.71} & 67.47$\pm$2.14 & \underline{65.06$\pm$1.84} & \underline{64.64$\pm$1.57} & \underline{66.42$\pm$1.31} & \underline{59.37} \\
 & MAGPrompt & 32.22$\pm$1.73 & \underline{68.20$\pm$1.98} & 61.79$\pm$3.74 & 62.58$\pm$1.96 & 66.33$\pm$1.73 & 58.22 \\
 & MAGPrompt+ & \textbf{36.39$\pm$2.09} & \textbf{69.00$\pm$1.41} & \textbf{65.58$\pm$1.15} & \textbf{65.41$\pm$1.83} & \textbf{68.06$\pm$1.93} & \textbf{60.89} \\
\bottomrule
\end{tabular}
\end{adjustbox}
\end{table*}

\begin{table*}
\centering
\caption{Average accuracy on full-shot graph classification tasks over MoleculeNet datasets. Best results are in bold; second-best are underlined.}
\label{tab:molecule_full}
\begin{adjustbox}{width=0.85\textwidth}
\begin{tabular}{l|c|cccc|c}
\toprule
Pre-training & Tuning & BACE & BBBP & ClinTox & SIDER & Avg. \\
\midrule
 & Fine-tune & 86.58$\pm$4.56 & 91.77$\pm$4.10 & 64.99$\pm$13.10 & 60.05$\pm$4.40 & 75.85 \\
 & GPF & 83.69$\pm$5.51 & 92.73$\pm$2.82 & 68.28$\pm$12.89 & 60.28$\pm$5.36 & 76.25 \\
 & GPF-plus & 84.12$\pm$5.14 & 93.19$\pm$2.06 & 68.28$\pm$12.80 & 60.75$\pm$4.72 & 76.59 \\
AttrMasking & EdgePrompt & 85.51$\pm$5.71 & 92.14$\pm$2.98 & 64.92$\pm$16.18 & 60.16$\pm$5.06 & 75.68 \\
 & EdgePrompt+ & 86.06$\pm$5.54 & 92.79$\pm$3.57 & 66.06$\pm$15.17 & 60.30$\pm$4.97 & 76.30 \\
 & MAGPrompt & \underline{86.79$\pm$4.95} & \underline{93.20$\pm$2.18} & \underline{68.45$\pm$14.39} & \underline{60.84$\pm$5.39} & \underline{77.32} \\
 & MAGPrompt+ & \textbf{86.82$\pm$5.21} & \textbf{94.32$\pm$2.38} & \textbf{71.81$\pm$11.58} & \textbf{62.01$\pm$4.36} & \textbf{78.74} \\

 \midrule
 
 & Fine-tune & \textbf{86.49$\pm$2.99} & 93.46$\pm$4.47 & 65.33$\pm$15.52 & 59.42$\pm$5.97 & 76.17 \\
 & GPF & 84.02$\pm$4.21 & \underline{94.02$\pm$2.20} & \underline{70.25$\pm$17.68} & 59.01$\pm$6.17 & \underline{76.83} \\
 & GPF-plus & 83.69$\pm$5.36 & 91.04$\pm$4.12 & \textbf{70.83$\pm$15.06} & 59.95$\pm$6.26 & 76.38 \\
DGI & EdgePrompt & 83.36$\pm$4.56 & 93.17$\pm$3.20 & 61.53$\pm$15.41 & 60.09$\pm$5.82 & 74.54 \\
 & EdgePrompt+ & 83.74$\pm$4.42 & 92.55$\pm$2.41 & 60.77$\pm$18.36 & 59.21$\pm$4.30 & 74.07 \\
 & MAGPrompt & \underline{85.10$\pm$5.00} & 93.63$\pm$3.32 & 67.43$\pm$14.18 & \underline{60.59$\pm$4.66} & 76.69 \\
 & MAGPrompt+ & 84.97$\pm$4.82 & \textbf{94.23$\pm$2.69} & 67.75$\pm$13.07 & \textbf{61.71$\pm$4.69} & \textbf{77.16} \\

 \midrule
 
 & Fine-tune & 83.80$\pm$6.09 & 92.68$\pm$3.70 & 62.20$\pm$11.50 & \underline{61.50$\pm$4.84} & 75.05 \\
 & GPF & 82.92$\pm$5.55 & 92.35$\pm$3.44 & \textbf{73.47$\pm$10.74} & 60.06$\pm$3.08 & \underline{77.20} \\
 & GPF-plus & 83.82$\pm$5.07 & 92.78$\pm$2.91 & 70.03$\pm$14.56 & 60.33$\pm$4.17 & 76.74 \\
ContextPred & EdgePrompt & 83.23$\pm$6.02 & 92.74$\pm$3.44 & 65.36$\pm$16.03 & 60.34$\pm$5.01 & 75.42 \\
 & EdgePrompt+ & 83.51$\pm$4.83 & 92.66$\pm$3.98 & 67.91$\pm$12.73 & 59.90$\pm$2.73 & 76.00 \\
 & MAGPrompt & \underline{84.91$\pm$5.13} & \underline{93.41$\pm$3.48} & 69.78$\pm$10.83 & 60.39$\pm$4.17 & 77.12 \\
 & MAGPrompt+ & \textbf{84.99$\pm$4.67} & \textbf{94.29$\pm$2.81} & \underline{70.65$\pm$13.61} & \textbf{61.74$\pm$3.89} & \textbf{77.92} \\

 \midrule
 
 & Fine-tune & 84.67$\pm$4.76 & \underline{92.99$\pm$3.71} & 64.55$\pm$9.48 & \textbf{62.17$\pm$4.27} & 76.09 \\
 & GPF & 83.00$\pm$5.91 & 91.72$\pm$3.71 & 66.11$\pm$9.21 & 59.12$\pm$6.61 & 74.99 \\
 & GPF-plus & 85.10$\pm$5.32 & 91.93$\pm$3.52 & \underline{66.60$\pm$6.39} & 60.05$\pm$5.39 & 75.92 \\
EdgePred & EdgePrompt & 84.32$\pm$5.77 & 92.60$\pm$3.02 & 63.06$\pm$10.81 & 60.10$\pm$6.90 & 75.02 \\
 & EdgePrompt+ & 85.00$\pm$5.58 & 92.68$\pm$3.14 & 65.15$\pm$13.44 & 60.76$\pm$4.95 & 75.90 \\
 & MAGPrompt & \underline{85.66$\pm$5.27} & 92.45$\pm$3.11 & 65.34$\pm$10.61 & 61.23$\pm$6.00 & \underline{76.17} \\
 & MAGPrompt+ & \textbf{86.18$\pm$4.22} & \textbf{93.48$\pm$2.62} & \textbf{66.90$\pm$11.95} & \underline{61.26$\pm$6.12} & \textbf{76.96} \\
\bottomrule
\end{tabular}
\end{adjustbox}
\end{table*}

\subsection{Analysis of Prompt Collapse}
\label{subsec:prompt_collapse}

\begin{figure*}[!ht]
    \centering
    \includegraphics[width=\linewidth]{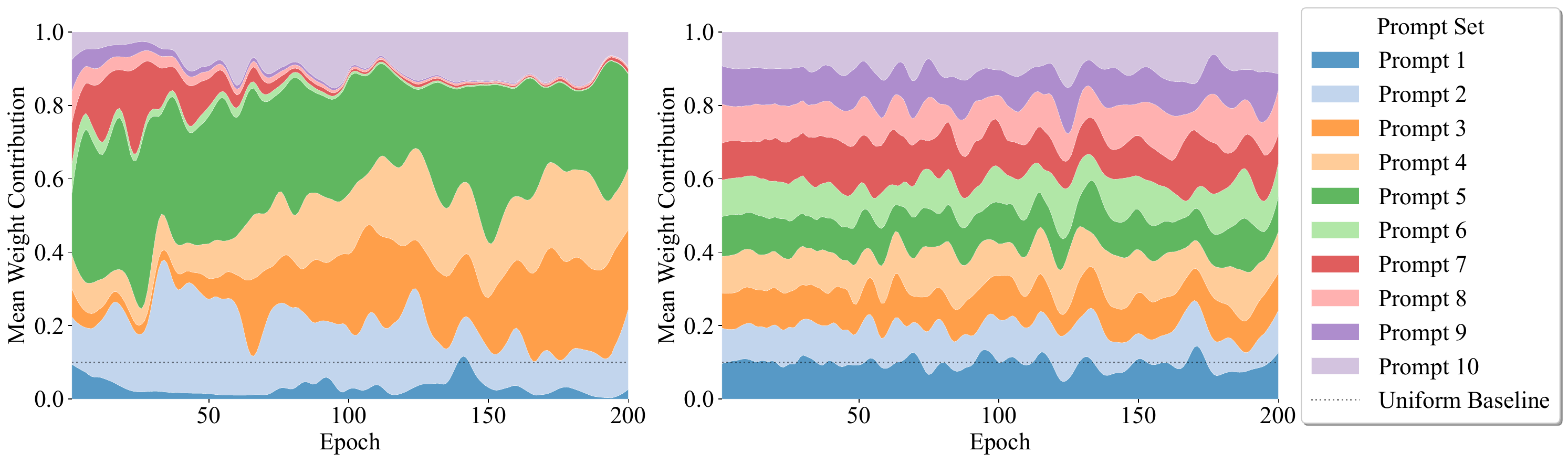}
    \caption{
    Evolution of prompt mixture weights during training \textbf{without} (left) and \textbf{with} (right) the prompt-collapse regularization $\mathcal{L}_{\mathrm{pc}}$.
    Each stacked area shows the mean contribution of a prompt basis across epochs, while the dotted line indicates the uniform baseline.
    Without $\mathcal{L}_{\mathrm{pc}}$, the prompt distribution collapses toward a small subset of prompts, whereas $\mathcal{L}_{\mathrm{pc}}$ encourages balanced and stable utilization of all prompt bases.
    }
    \label{fig:prompt_collapse}
\end{figure*}

To better understand the role of the prompt-collapse regularization $\mathcal{L}_{\mathrm{pc}}$, we analyze the training dynamics of prompt mixture weights in MAGPrompt+.
Figure~\ref{fig:prompt_collapse} visualizes the mean contribution of each prompt basis across training epochs, with the uniform distribution shown as a reference.

Without $\mathcal{L}_{\mathrm{pc}}$ (left), the model quickly exhibits prompt collapse, where the mixture weights concentrate on a small subset of prompts while others receive negligible contribution.
This behavior reduces the effective capacity of the prompt set and limits the expressiveness of message-adaptive prompting.
In contrast, incorporating $\mathcal{L}_{\mathrm{pc}}$ (right) encourages balanced and stable utilization of multiple prompt bases throughout training, preventing collapse and preserving prompt diversity.

These observations explain the consistent performance gains of MAGPrompt+ observed in Table~\ref{tab:each_component}, and highlight the importance of $\mathcal{L}_{\mathrm{pc}}$ for stabilizing optimization when employing multiple prompt bases.

\section{Limitations and Future Work}
The effectiveness of MAGPrompt depends on the presence of neighbor contribution mismatch between pre-training and downstream tasks; when pre-trained aggregation patterns already align well with downstream objectives, performance gains may be limited. Although MAGPrompt is {parameter-efficient} compared to full fine-tuning, the additional edge-wise gating introduces modest computational overhead, which may be unnecessary for shallow architectures or small graphs. Moreover, our approach focuses on adapting local message passing and does not explicitly incorporate global or higher-order structural information, potentially limiting its effectiveness on tasks dominated by long-range dependencies. Finally, our evaluation is restricted to standard message-passing GNN backbones; extending message-adaptive prompting to alternative graph architectures and developing adaptive mechanisms for prompt capacity and computational efficiency are promising directions for future work.

\end{document}